\documentclass{article}

 \usepackage[preprint]{neurips_2026}

\usepackage[utf8]{inputenc} 
\usepackage[T1]{fontenc}    
\usepackage{hyperref}       
\usepackage{url}            
\usepackage{booktabs}       
\usepackage{amsfonts}       
\usepackage{amsmath}        
\usepackage{amsthm}         
\usepackage{nicefrac}       
\usepackage{microtype}      
\usepackage[dvipsnames]{xcolor}         
\usepackage{graphicx}

\newcommand{\doiurlfoot}[1]{\footnote{\url{https://doi.org/#1}}}
\newcommand{\sourceurlfoot}[1]{\footnote{\url{#1}}}
\newtheorem{proposition}{Proposition}

\title{Concurrence of Symmetry Breaking and Nonlocality Phase Transitions in Diffusion Models}

\author{%
  Yifan F.~Zhang \\
  Department of Electrical and Computer Engineering\\
  Princeton University\\
  Princeton, NJ 08544, USA \\
  \texttt{yz4281@princeton.edu} \\
  \And
  Fangjun Hu \\
  QuEra Computing Inc.\\
  1284 Soldiers Field Road, Boston, MA 02135, USA \\
  \texttt{fhu@quera.com} \\
  \And
  Guangkuo Liu \\
  JILA and Department of Physics\\
  University of Colorado Boulder\\
  Boulder, CO 80309, USA \\
  \texttt{guangkuo.liu@colorado.edu} \\
  \And
  Mert Okyay \\
  CTQM and Department of Physics\\
  University of Colorado Boulder\\
  Boulder, CO 80309, USA \\
  \texttt{mert.okyay@colorado.edu} \\
  \And
  Xun Gao \\
  JILA and Department of Physics\\
  University of Colorado Boulder\\
  Boulder, CO 80309, USA \\
  \texttt{xun.gao@colorado.edu} \\
}

\begin{document}

\maketitle

\begin{abstract}
Diffusion models undergo a phase transition in a critical time window during generation dynamics, with two complementary diagnoses of criticality. The symmetry breaking picture views the critical window as when trajectories bifurcate into different semantic minima of the energy landscape, whereas the nonlocality picture views the critical window as when local denoising fails.
We study whether two notions of such phase transitions are concurrent in modern diffusion transformers. By evaluating the dynamics and outcomes of the generation trajectory, we observe a near-simultaneous occurrence of the non-locality and symmetry breaking critical times.
%
%
%
%
Our work is the first to unify the two notions of phase transitions in practice: it provides a concrete diagnostic for when and why diffusion models rely on conditioning and global denoising, enabling principled evaluation of model efficiency and guiding the design of architectures and sampling schemes that avoid unnecessary computation.

\end{abstract}

\begin{center}
\noindent\textbf{Code:} \url{https://github.com/zsxqblz/symmetry_nonlocality_transition}
\end{center}

\section{Introduction}

Diffusion models now produce high-quality images, but we still have a limited understanding of \emph{when} different capabilities turn on during generation. A diffusion trajectory is not a uniform process: the model does not use all signals equally at all times. One influential line of work borrows techniques from statistical mechanics and interprets the generation process as a \emph{symmetry breaking} \footnote{Although ``symmetry breaking" is a somewhat imprecise term—--given that semantic minima are not necessarily symmetric--—we adopt it because it is standard in the literature \cite{raya2023spontaneous} and effectively conveys the underlying physical concept.} phase transition, where the distribution evolves from a single unstructured basin to multiple class- or prompt-consistent basins~\cite{raya2023spontaneous}. This picture predicts the existence of a ``critical window'' of time during which the dynamics has drastic impact on the output, such as changing the class or prompt semantics, while similar interventions outside that window have much smaller effects.



The statistical mechanics perspective is intuitive (and natural), as the diffusion model essentially mimics a non-equilibrium ``cooling'' from high-temperature Gaussian noise to low-temperature data distribution~\cite{sohl-dickstein-diffusion}.  However, this symmetry breaking paradigm is ignorant of the spatial structure of the data. For example, the image distribution undergoes the same symmetry breaking transition regardless of whether the model treats the image as a vectorization of pixels or as a structured 2D array. Since locality bias is a key inductive bias in dataset \cite{kamb2024analytic} and model design, it is important to understand how it interacts with the phase-transition picture.



Tools and concepts to tackle the connection between phase transitions and locality has recently been developed in the many-body physics and quantum error correction community: a complementary notion of  criticality, a \emph{nonlocality} phase transition, has been proposed \cite{sang_mixed_phase}. This perspective puts locality front and center, and asks when local neighborhoods of a site contains enough information for denoising. At a higher noise rate, information is further corrupted, so the denoiser needs to look at a larger neighborhood to recover one patch. The noise level at which the denoising radius diverges is understood as a phase transition in which global information is needed to perform denoising, which may contain information such as the symmetry sector in which the initial condition was. In diffusion models, the identical information-theoretic probe has been found to gives rise to a critical window in which local denoising fails~\cite{hu2025localdiffusionmodelsphases}. This criticality may be inherently distinct from phase transitions in equilibrium statistical mechanics, e.g., because it is based on a purely operational criterion and does not follow from a free-energy argument.


This paper aims to reconcile these two notions in frontier models. Our starting hypothesis is simple: they should be related because semantics typically carry nonlocal information. If a model is actively injecting semantic identity (for example, deciding between a Golden Retriever dog and a golden British Shorthair cat, which can share similar local fur textures but differ in global shape and anatomy), it should also require long-range context. Conversely, if denoising can be done from local texture statistics alone, strong semantic guidance should be less necessary. In short, these arguments suggest that \emph{symmetry breaking predicts when semantic forcing is needed, and nonlocality predicts when nonlocal computation is needed}; if semantics are globally organized, the two windows should align.

\begin{figure}
  \centering
  \includegraphics[width=\linewidth]{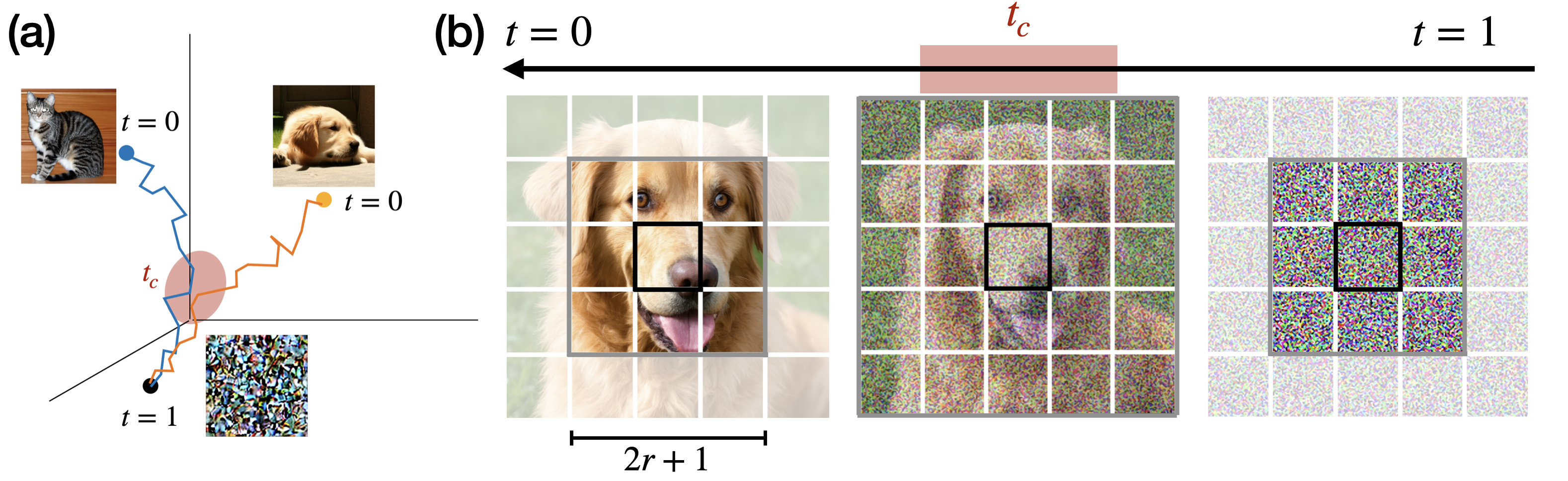}
  \caption{\textbf{Two definitions of phase transitions}. (a) Symmetry breaking phase transition. At $t=1$, the energy landscape has one global minimum. As $t$ approaches the phase transition, multiple local minima start to appear. (b) Nonlocality phase transition. Inside a phase, noise on a patch (black box) can be removed by using information in a small neighborhood (gray box). Near the phase transition, this neighborhood grows to be comparable to the entire image.}
  \label{fig:setup}
\end{figure}

Figure~\ref{fig:setup} clarifies the two notions of phase-transitions. Panel (a) depicts the critical window for symmetry breaking: as diffusion proceeds, trajectories in high-dimensional state space branch into distinct semantic basins (for example, cat-like versus dog-like minima). Panel (b) depicts a locality transition for patch denoising: once global semantic identity is already stable, recovering a local region (such as a dog's nose) may require only a small neighborhood (such as the head), whereas near a critical window where semantic identity is still unresolved or ambiguous, denoising must use broader, often global, context to inject the correct semantics.

\subsection{Summary of Main Results}

We probe the two definitions of critical window numerically from modern diffusion transformers (DiTs) and examine how they align. 
We use two complementary families of probes. The \emph{instantaneous probe} directly measures score vectors along different trajectories. It compares conditional versus unconditional scores, and measures the symmetry breaking critical window where the difference is large. It also compares the global versus locally-approximated scores, and measures the nonlocality critical window where the difference is large. If the two critical windows are aligned, we conclude that the two phase transitions occur at the same time.

We complement the instantaneous probe with a series of \emph{integrated probes}, which focus on how the dynamics influence the final generated samples. These probes include the standard forward-backward experiments~\cite{dynamical_regimes,li2025blink} used to define symmetry breaking critical times, as well as a series of sliding window methods where we switch between conditional and unconditional, global and local denoisers in a time window of the denoising process. We benchmark the quality of generation and ask when global or conditional denoising is necessary to complete the tasks with high quality.

We find concurrence of symmetry breaking and nonlocality in both probes. For instantenous probes in the Facebook DiT-XL model \cite{peebles2023scalable}, the conditional versus unconditional gap becomes large in an early window centered near $t\approx 0.2$, and the local versus global gap is large in the same window. In Stable Diffusion 3 (SD3) Medium model \cite{esser2024scalingrectifiedflowtransformers}, both gaps concentrate very close to the noisy endpoint $t=1$. The co-occurrence of the two gaps gives operational evidence that semantic forcing and nonlocal computation turn on together.

In integrated probes, we find concurrence as well. Facebook DiT-XL shows a critical window for both symmetry breaking and nonlocality near $t\approx 0.5$. This window is later than the window found in instantaneous probes, which can be interpreted as suboptimality: the model uses conditioning and nonlocal computation at an earlier $t$ than when it is actually needed.  In contrast, the higher-performance model, SD3 medium, shows a concurrent locality and symmetry breaking phase transition close to $t=1$ in both instantaneous probes and integrated probes.

\begin{figure}
  \centering
  \includegraphics[width=\linewidth]{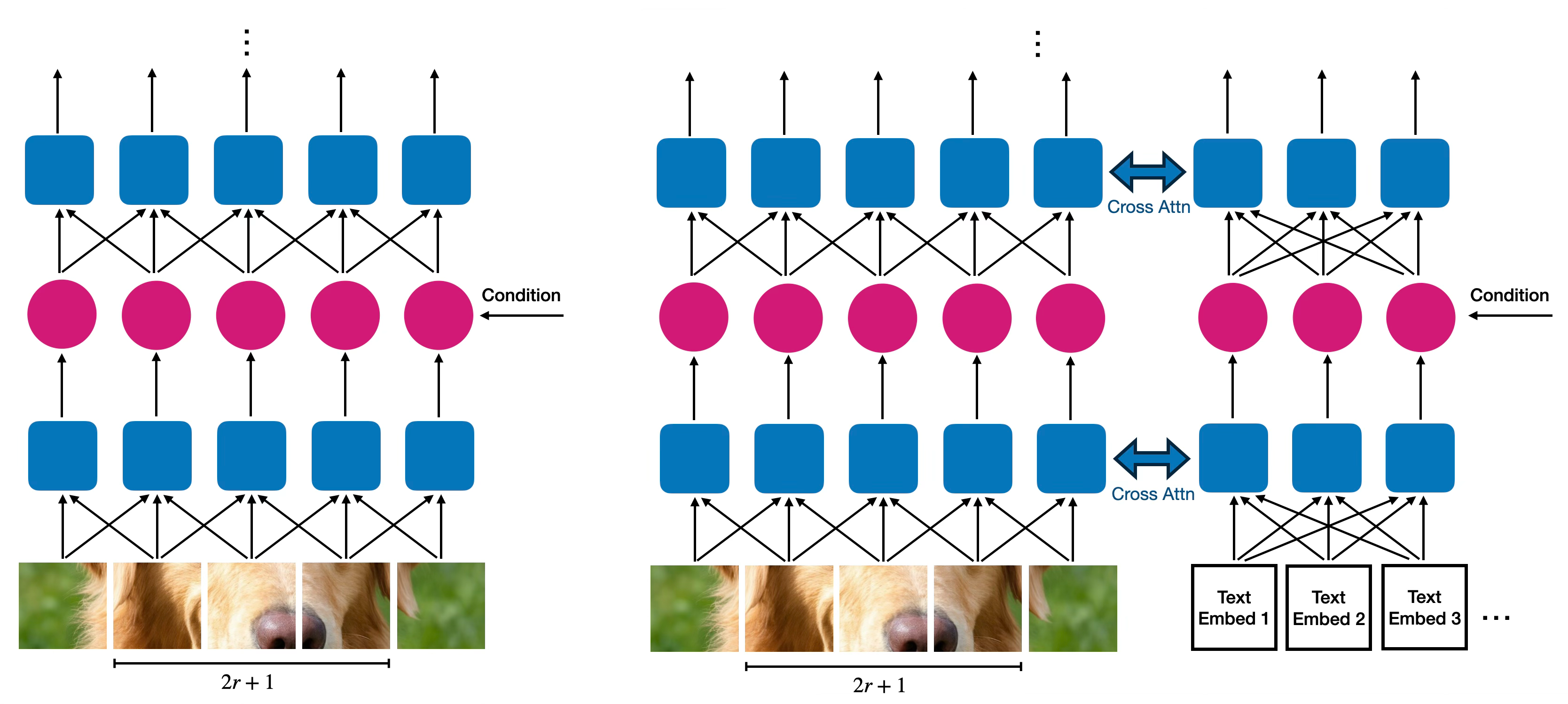}
  \caption{\textbf{DiT with local attention}. (a) In class-conditioned DiT, we restrict image-token attention to a local window with radius $r$ and window size $2r+1$. We visualize a one-dimensional strip of image tokens for simplicity. The conditioning signal still affects each token. (b) In multi-model DiT (MMDiT), we restrict image token--image token attention to a local window with radius $r$. Text token--text token and image token--text token attentions remain dense, so conditioning enters through both token-wise modules and cross attention.}
  \label{fig:local-attention}
\end{figure}

\section{Theoretical Background}

\subsection{Diffusion models and the score function}

Let $x_0 \sim p_{\mathrm{data}}$ denote a clean data sample (e.g., an image in pixel or latent space), and let $x_t$ denote its noised version at diffusion time $t \in (0,1]$ such that $t=1$ is the high-noise limit. 
The continuous-time Gaussian forward noising process can be written as an SDE
\begin{equation}
 \mathrm{d}x_t = f(x_t,t)\,\mathrm{d}t + g(t)\,\mathrm{d}W_t,
\end{equation}
whose time-reversal admits the reverse-time SDE \cite{DDPM20}
\begin{equation}
 \mathrm{d}x_t = \bigl[f(x_t,t) - g^2(t)\,s^*(x_t,t)\bigr] \mathrm{d}t + g(t)\,\mathrm{d}\bar W_t,
\end{equation}
where $\bar W_t$ is a standard Brownian motion running backwards in time. The function  $s^*(x_t,t)$ is called the score at time $t$, and it is the gradient of the log density $s^*(x_t,t) = \nabla_{x_t} \log q_t(x_t)$
where $q_t$ is the marginal distribution of $x_t$. Diffusion models learn a parametric approximation $s_\theta(x_t,t)$ to $s^*(x_t,t)$ (or an equivalent parameterization) via denoising score matching \cite{scoreSDE2021}. The learned score $s_\theta$ defines reverse-time dynamics that turns noise into a sample from the model distribution by initializing $x_1\sim\mathcal{N}(0,I)$ and numerically integrates the reverse SDE from $t=1$ to $t=0$. 

\subsection{Symmetry breaking in sampling and classifier-free guidance} 

Its origin in statistical physics invites a study of diffusion dynamics, where early work noticed varying relative importance of different times~\cite{ddim2022,perceptual_imp2022,raya2023spontaneous} for sample quality and class. These observations were studied in physics-inspired toy models~\cite{Birolicurieweiss_2023, Ambrogioni_stochastic_2025}, motivating the terminology `symmetry-breaking' for when the output sample is placed in a class during the diffusion path.

In many applications we want to sample the conditional distribution $p_\theta(x \mid y)$ for a specified condition $y$ (in addition to the marginal $p_\theta(x)$), which requires access to the conditional score $\nabla_{x_t} \log p_\theta(x_t \mid y)$ for all $t$. The modern approach to obtain a conditional sampler is known as classifier-free guidance (CFG)~\cite{ho2022classifier}, where the \emph{same} parameters learn both a conditional score $s_\theta(x_t,t\mid y)$ and an unconditional score $s_\theta(x_t,t)$. During sampling, CFG mixes the two scores to form a `guided score'
\begin{equation}
 s_{\mathrm{cfg}}(x_t,t\mid y) 
 = s_\theta(x_t,t) + w\bigl(s_\theta(x_t,t\mid y) - s_\theta(x_t,t)\bigr),
\end{equation}
where $w\ge 0$ is the guidance scale. To make contact with physics, we write scores as gradients of energies $U$ such that $s= -\nabla U$, and CFG can be viewed as sampling from an energy tilted by an external field:
\begin{equation}
 U_{\mathrm{cfg}}(x_t,t\mid y) = (1-w)U_{\mathrm{uncond}}(x_t,t) + w\,U_{\mathrm{cond}}(x_t,t\mid y) + \text{const}.
\end{equation}
Viewing each class as as a local minima of the conditional energy landscape $U_{\mathrm{cond}}(x_t,t\mid y)$, the CFG weighted term plays the role of a bias that breaks the symmetry. The absence of a symmetry-breaking field implies that there needs to be spontaneous symmetry breaking for the model to pick a class, which has been associated with experimental failures of forward-backward experiments~\cite{ambrogioni2026outofequilibriumphasetransitionsseed, hierarchical_diffusion}.

\subsection{Local score function via approximate Markovianity}

A seperate line of work characterises phases of quantum distributions by local recoverability: between two distributions~\cite{sang_mixed_phase} proposed the decay length-scale of conditional mutual information (CMI) (the `Markov length') to diagnose when one is not locally recoverable from the other. Bringing these insights to learning, \cite{hu_2026_learning} has  shown that if the CMI remains short-ranged during diffusion, then one can construct a sampler that outputs from this distribution using only local operations. On the other hand, \cite{kumar_2026_unlearnable} shows that distributions with long-ranged CMI is generically hard to learn.

We encode locality via a tripartition of the image $X=(x_A,x_B,x_C)$ as shown in Figure~\ref{fig:setup} where $A$,$B$, $C$ denote the innermost box, the surrounding annulus, and the rest, respectively. If the distribution $p(x_A,x_B,x_C)$ forms a Markov chain, i.e., $
p(x_A\mid x_B,x_C)=p(x_A\mid x_B)$,
then the score on region $A$ depends only on $(x_A,x_B)$ where $ 
\nabla_{x_A}\log p(x_A,x_B,x_C) = \nabla_{x_A}\log p(x_A\mid x_B)$ and the CMI would vanish. This property might hold approximately, which is quantified as rapidly-decaying CMI
\begin{equation}
 I\bigl(A:C\mid B\bigr) \approx I_0\,\exp\!\left(-d_{AC}/\xi\right),
\end{equation}
where $d_{AC}$ is a distance between regions $A$ and $C$, and $\xi$ is the Markov length. In such cases, we can find a good local approximation to the true score $s_A(x_A,x_B, x_C)$, where $\smash{s^{\mathrm{loc}}_A(x_A,x_B) = \nabla_{x_A}\log p(x_A\mid x_B)}$,
which we expect to be accurate in real images. For example, to generate the nose of a dog, the model may only need to look at a neighborhood containing the snout and eyes, and not the rest of the image.

\section{Experiments}

We study two diffusion transformer systems: a Facebook DiT-XL class-conditional ImageNet model and SD3 medium, a text-conditioned MMDiT model. As before, throughout the experiments, $t=1$ denotes the noisy endpoint and $t=0$ denotes the clean endpoint.

\subsection{Constructing local denoisers}
Training a local denoiser could be computationally expensive. Instead, we take any pre-trained DiT and construct a local version of it, which also has the benefit of understanding model's internal circuitry. For each image token, we truncate the attention window to a small radius $r$ around itself as shown in Figure~\ref{fig:local-attention}(a). Note that for Facebook DiT-XL, class and time conditioning is injected into each patch independently through AdaLN~\cite{perez2018film,peebles2023scalable}. Since the conditioning is still computed globally, in principle the model is capable of providing global conditioning information to each patch even when image-token attention is local. For MMDiT model such as SD3, both image and text tokens enter the joint attention. Therefore, we only restrict image token--image token attention to a local window, while keeping text token--text token and image token--text token attentions dense, as described in Figure~\ref{fig:local-attention}(b). This way, the conditioning signal can still be globally available to each patch through both cross attention and text token self attention. 

We also note that the truncation procedure does not produce a strictly local denoiser, as context from patches further than $r$ can still enter by propagating through multiple local attention steps. One should therefore view the truncated DiT as an approximation with strong locality bias rather than a strictly local denoiser. This is in fact operationally meaningful, since we are performing a minimal perturbation to the original model architecture, which in principle saves a large amount of compute.



\subsection{Instantaneous probe: score gap}

We first probe the phase transition through the instantaneous probe, which measures the behavior of the score function along the sampling and training trajectories. We take the conditional score function $s_\theta(x_t,t \mid y)$ and the unconditional score function $s_\theta(x_t,t)$, define the \emph{conditioning gap} as the $l_2$ norm of their difference: 
\begin{equation}\label{eq:conditioning_gap}
  \Delta \vec{s}_{\mathrm{cond}}(x_t,t \mid y )
  = s_\theta(x_t,t)-s_\theta(x_t,t \mid y),
\end{equation}
The conditioning gap provides a direct, model-internal readout of how strongly the condition changes the denoising vector at time $t$.

For nonlocality, we construct local denoiser from pretrained model following Figure \ref{fig:local-attention} which provides a local score function $s_{\theta,r}(x_t,t)$ from the global score function $s_{\theta}(x_t,t)$. The \emph{locality gap} is then defined as the $l_2$ norm of their difference.

The truncated DiT gives a local approximation of the score function $s_{\mathrm{local},r}(x_t,t)$. to the original global score function $s_{\mathrm{global}}(x_t,t)$. We define the locality gap
\begin{equation}\label{eq:locality_gap}
  \Delta \vec{s}_{\mathrm{loc}}(x_t,t,r)
  = s_{\theta}(x_t,t)-s_{\theta,r}(x_t,t).
\end{equation}
The above definition is valid for both conditional and unconditional score function. If the trained model can compute the relevant score locally at time $t$, this gap should be small for modest $r$. If global information is necessary, the gap should remain large until $r$ is sufficiently large. This gives a direct operational version of the CMI-based nonlocality probe. Throughout this paper, we sample images of $256\times 256$ pixels for Facebook DiT and  $512\times 512$ pixels for SD3 medium. Both models we consider have patch size of 16 pixels, so in our study, setting $r=16$ for Facebook DiT / $r=32$ for SD3 medium reduces local denoiser to global denoiser.

\begin{figure}
  \centering
  \includegraphics[width=\linewidth]{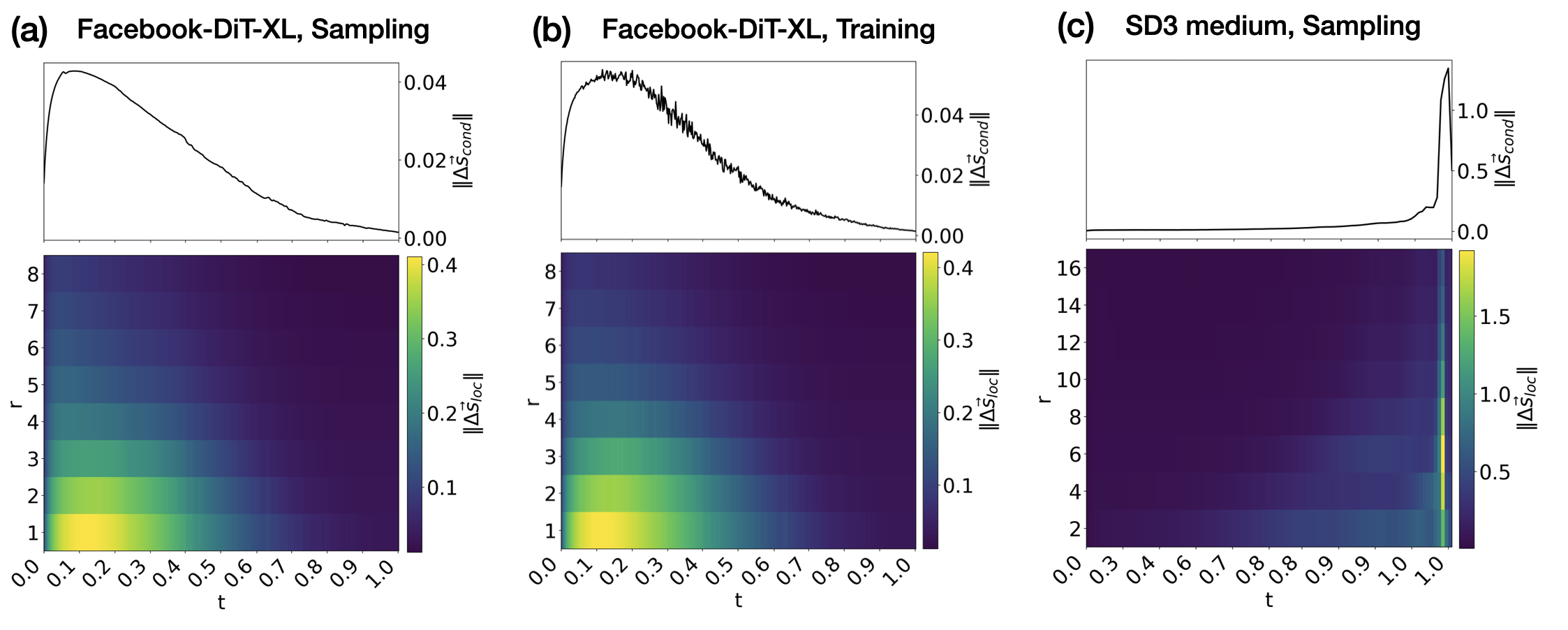}
  \caption{\textbf{Conditioning gap vs. locality gap}. Top row: the one-dimensional plot shows the conditioning gap $\|\Delta \vec{s}_{\mathrm{cond}}\|$ as a function of $t$. Bottom row: the heatmap shows the locality gap $\|\Delta \vec{s}_{\mathrm{loc}}\|$, defined using conditional score function, as a function of attention radius $r$ and time $t$. (a) Facebook DiT-XL along the sampling trajectory. (b) Facebook DiT-XL along the training trajectory. (c) SD3 medium along the sampling trajectory; the horizontal axis is reversed so that sampling proceeds left to right. In all three panels, the two gaps align remarkably well despite being functionally independent probes.}
  \label{fig:score-gap}
\end{figure}

We first ask whether the two score gaps identify the same time window. Figure~\ref{fig:score-gap} plots the conditioning gap on the top row and the locality gap on the bottom row. We normalize the gap by the number of image tokens. The top row measures the strength of the semantic field induced by conditioning. The bottom row measures the cost of replacing the global denoiser with a local one: large values indicate that the model's score cannot be reproduced from local image context alone.

For Facebook DiT-XL, Figure~\ref{fig:score-gap}(a,b), the conditioning gap is initially small at $t$ close to zero. This is intuitive, as the semantics are still present in slightly noisy images, so conditioning is unnecessary. As $t$ increases, it become large at saturate near $t\approx 0.2$. They it decays as the trajectory approaches the clean endpoint. This behavior is consistent with the phenomenology of symmetry breaking phase transition, and is also consistent with the empirical observation of when conditioning is beneficial~\cite{kynkaanniemi2024applying,jin2025stage}.

The behavior of the locality gap is largely aligned with the conditioning gap. at early and late time, locality gap quickly decays at increasing $r$, indicating that local denoisers approximate global denoiser well. Around $t \approx 0.2$, locality gap remains large even at large $r$. The behaviors are similar between training and sampling trajectories, indiciating a good generalization of the model.

For SD3 medium, Figure~\ref{fig:score-gap}(c,d), the behavior of both gaps are large the same and are aligned. We also observe a richer behavior when sending different prompts. For example, when prompting for violin (Figure~\ref{fig:score-gap}(d)), we observe two peaks in conditioning and locality gaps near $t=1$. We also observe a weaker peak at a lower $t$.



The conditioning and locality gaps are independent quantities: one compares conditional and unconditional scores, while the other compares global and local computation under the same conditioning setup. Their alignment therefore gives direct evidence that the symmetry breaking transition and the nonlocality transition are linked. The locality-gap heatmaps play the same role operationally that CMI heatmaps play in~\cite{hu2025localdiffusionmodelsphases}: both identify times when distant context is needed, but our probe uses an actual truncated denoiser rather than an information-theoretic decoder.

\subsection{Integrated probe: forward-backward error correction}

\begin{figure}
  \centering
  \includegraphics[width=\linewidth]{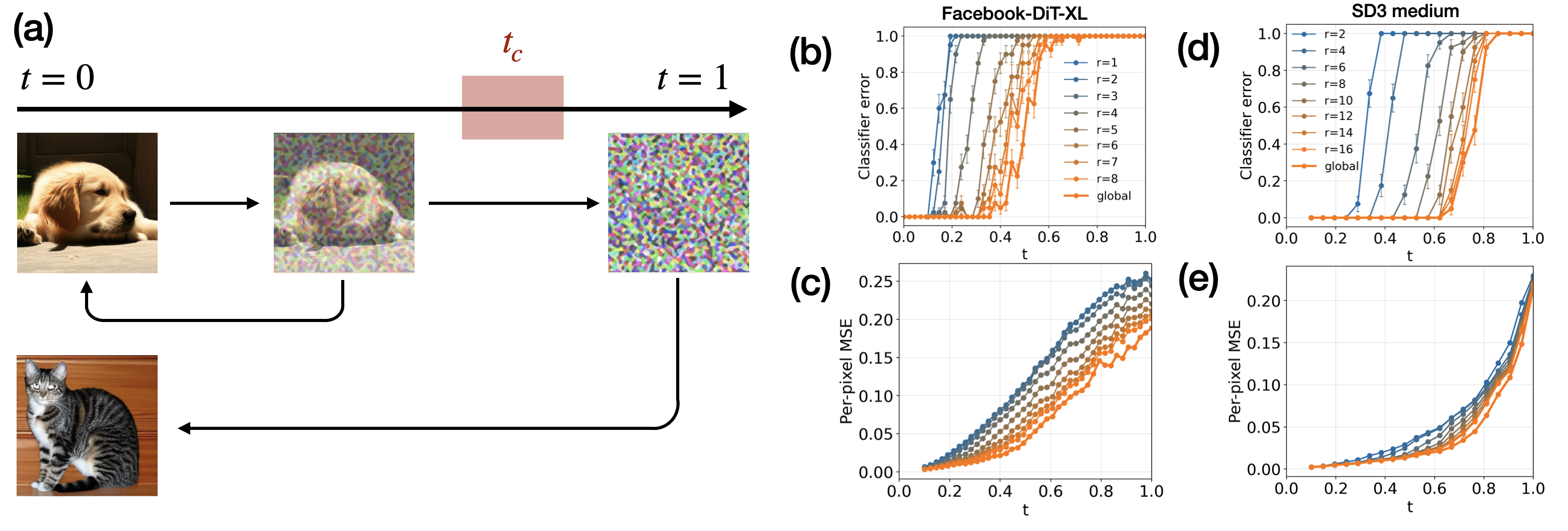}
  \caption{\textbf{Forward-backward experiment}. (a) Schematic of the error-correction experiment. When the noise level is below a threshold, the unconditional denoiser recovers the same class; when the noise level is above the threshold, the unconditional denoiser can recover a different class. (b,d) Classification error of images denoised from different noise levels $t$ for Facebook DiT-XL and SD3 medium. Curves labeled by $r$ use local denoisers with attention radius $r$ (see Fig~\ref{fig:local-attention}); the global curve uses the original denoiser. (c,e) Per-pixel MSE of the denoised image relative to the clean image.}
  \label{fig:partial-denoising}
\end{figure}

The model-internal details captured by gaps~\eqref{eq:conditioning_gap} and~\eqref{eq:locality_gap} may be smeared out during sampling (which integrates over time), so we need integrated probes which directly benchmark the downstream samples. The first is the `forward-backwards experiment' illustrated in Figure~\ref{fig:partial-denoising}(a): start from a clean image, add noise forward to time $t$, and then denoise backward unconditionally. If the noised state remains inside the basin of the original semantic class, the unconditional denoiser recovers an image from the same class. If $t$ crosses the semantic error threshold, the unconditional denoiser can return an image from a different class. This is also a standard probe of symmetry breaking phase transition in early studies by~\cite{dynamical_regimes,sclocchi2025phase,li2025blink}.

Figures~\ref{fig:partial-denoising}(b,d) plot the classifier-error rate of the denoised image using Facebook DiT-XL and SD3 medium. Both models show sharp classifier-error jumps as $t$ increases. For Facebook DiT-XL, Figure~\ref{fig:partial-denoising}(b), the global threshold is around $t\approx 0.5$. Local denoisers also exhibit sharp thresholds, and those thresholds move toward the global threshold as $r$ increases. This indicates that local denoising has its own error-correction transition, and that larger local neighborhoods recover more of the global denoiser's semantic capability. 

For SD3 medium, Figure~\ref{fig:partial-denoising}(d), the corresponding threshold is very close to $t=1$, and local thresholds again approach the global threshold as $r$ grows. It is expected that SD3 medium has a larger threshold: it is more powerful and therefore should be a better decoder. This leads to the following prediction: stronger models have critical window at larger $t$. We leave the validation of this prediction to future work.

Figures~\ref{fig:partial-denoising}(c,e) show a complementary low-level metric. The per-pixel MSE increases smoothly and sublinearly with $t$, with an almost flat derivative at small $t$. This is consistent with error correction: small forward noise can be removed without large reconstruction error. However, MSE does not expose the semantic transition. The semantic class can be destroyed abruptly while the pixelwise error curve remains smooth, because semantic information is nonlocal and is not visible in a per-pixel distortion measure.

This experiment also separates two kinds of alignment. Within the forward-backward probe, the global and local denoisers both show sharp thresholds, and the local thresholds converge toward the global threshold with larger radius. Across probes, however, Facebook DiT-XL is misaligned: the instantaneous score-gap window is near $t\approx 0.2$, while the integrated semantic error threshold is near $t\approx 0.5$. We interpret this as evidence that Facebook DiT-XL performs unnecessary conditioning and nonlocal computation at low $t$. SD3 medium is more aligned across probes, with close to $t=1$ transitions predicted by both instantaneous and integrated probes.

\subsection{Integrated probe: windowed conditioning}
We next ask when conditioning is necessary for actual sampling. The symmetry breaking picture predicts that conditioning is most important near the critical window. We first demonstrate this idea with Facebook DiT-XL in Figure~\ref{fig:condition-window-imagenet} Results for SD3 medium are in the appendix. Figure~\ref{fig:condition-window-imagenet}(a) shows that applying conditioning only during $[0.2,0.7]$, which covers the critical window predicted from forward-backward experiment, can produce a recognizable golden retriever. Figure~\ref{fig:condition-window-imagenet}(b) shows the complementary intervention: applying conditioning outside $[0.2,0.7]$ but removing it inside the window produces a much worse semantic result. Thus the outside-window conditioning is not enough, even though local textures and image statistics may still look plausible. 

To quantitatively pin down the critical window, we choose a short interval $[t_i,t_i+0.1]$ for conditioning and scan over $t_i$. We expect that when $[t_i,t_i+0.1]$ overlaps with the critical window the most, the sample has better quality. Figure~\ref{fig:condition-window-imagenet}(c) shows that classifier error is lowest around $t_i\approx 0.4$, and Figure~\ref{fig:condition-window-imagenet}(d) shows that Fréchet inception distance (FID) to the always-conditioned baseline is also lowest around the same time. The samples in Figure~\ref{fig:condition-window-imagenet}(e) visualize this dependence: semantic correctness emerges when conditioning is placed in the critical window. This indicates an alignment with the forward-backward experiment. The SD3 medium windowed-conditioning experiment is shown in the appendix (Figure~\ref{fig:condition-window-sd3}) and exhibits similar alignment.

We also note that the classifier error and FID are overall large. This is because we heavily corrupt the model and sampling dynamics. The argument is not to have a small classifier error / FID overall, but to test which $t_i$ corrupts these benchmarks minimally.

\begin{figure}
  \centering
  \includegraphics[width=\linewidth]{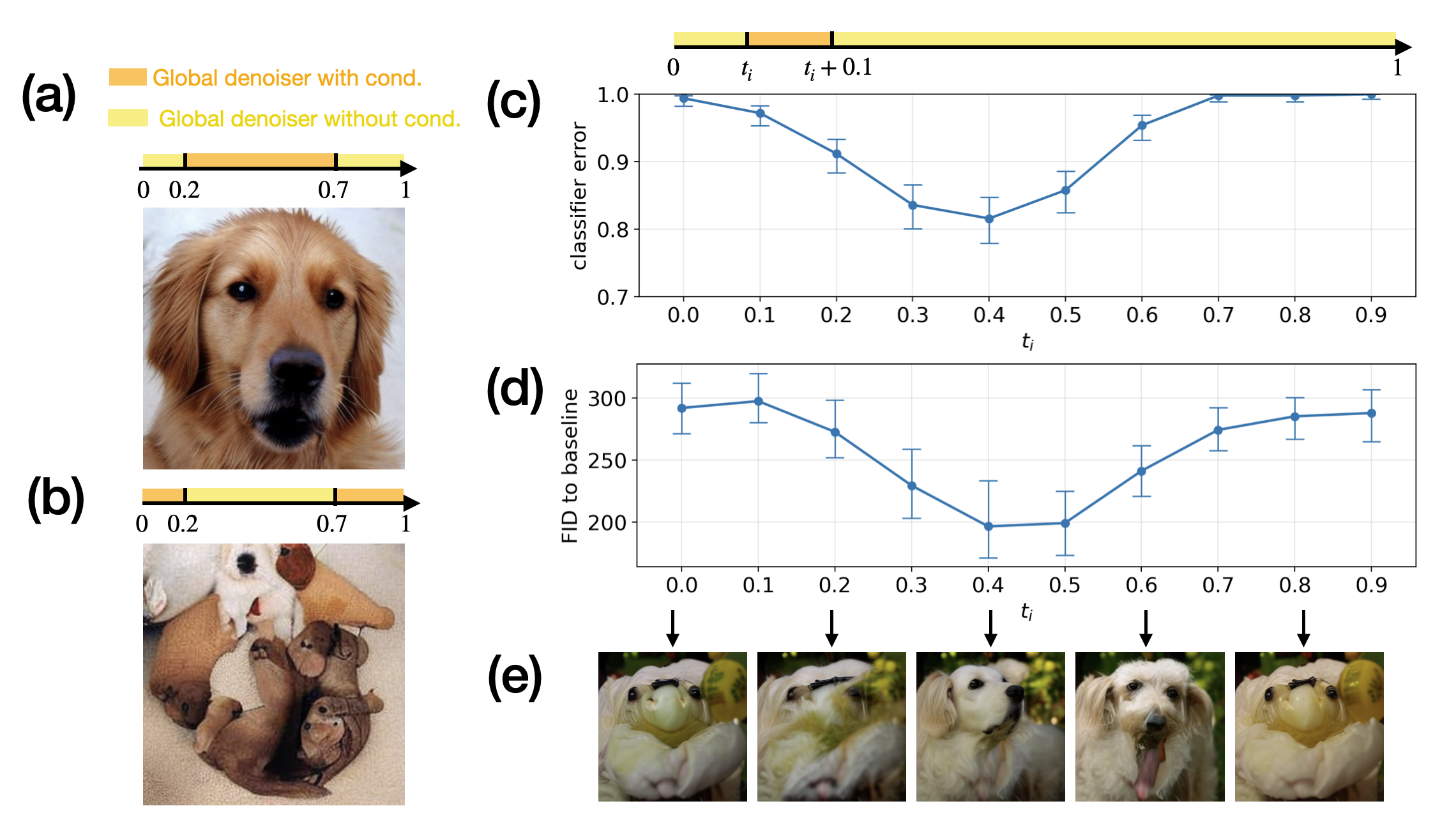}
  \caption{\textbf{Time-windowed conditioning in Facebook DiT-XL}. Left column: (a) Golden retriever sample generated with conditioning applied only in the time window $[0.2,0.7]$ and no conditioning outside the window. (b) Sample generated with conditioning removed in $[0.2,0.7]$ and applied outside the window. Right column: samples generated with conditioning applied only in short windows $[t_i,t_i+0.1]$ while scanning $t_i$. (c) Classification error as a function of $t_i$. (d) FID to always-conditioned samples as a function of $t_i$. (e) Visualization of generated samples at different $t_i$.}
  \label{fig:condition-window-imagenet}
\end{figure}


\subsection{Integrated probe: windowed local/global denoising}
The windowed-conditioning experiments test when semantic conditioning is needed. We now test when global denoising is needed. In these experiments, a global denoiser is used inside a time window and a local denoiser is used outside it, or vice versa.

We start with visual examples using Facebook DiT-XL (SD3 medium in Appendix). Figure~\ref{fig:locality-window-imagenet}(a)(i) compares the two extremes: an always-global denoiser produces a clean golden retriever, while an always-local denoiser gives ``domains'' of small dogs (with size comparable to $r$) smoothly merged together. This can be understood as the creative samples from~\cite{kamb2024analytic}. Figure~\ref{fig:locality-window-imagenet}(a)(ii) shows that using the global denoiser during $[0.2,0.7]$ and a local conditioned denoiser outside that window preserves a globally coherent dog better than the reverse schedule. Figure~\ref{fig:locality-window-imagenet}(a)(iii) makes the separation from conditioning sharper: even when the outside-window local denoiser is unconditional, putting global denoising in the critical window remains helpful, whereas replacing the critical window by local denoising is much worse. 

To pin down the critical window for nonlocality phase transition, we scan a global conditioning window $[t_i-0.2,t_i+0.2]$ and use local unconditional denoiser outside. Figures~\ref{fig:locality-window-imagenet}(b,c) show that classifier error and FID are again minimized when the global-denoising window is centered near $t_i\approx 0.4$. The samples in Figure~\ref{fig:locality-window-imagenet}(d) visualize the same window dependence. We also observe a slight increase in the minima at increasing $r$. We hypothesize that this is related to the increasing threshold at increasing $r$ in the forward-backward experiment (Figure~\ref{fig:partial-denoising}). At a fixed $r$ the observed critical window is earlier than the actual critical window, but the difference vanishes at increasing $r$.
Together with previous integrated probe results, we establish an alignment between symmetry breaking and nonlocality transition, through error correction and sampling.

\begin{figure}
  \centering
  \includegraphics[width=\linewidth]{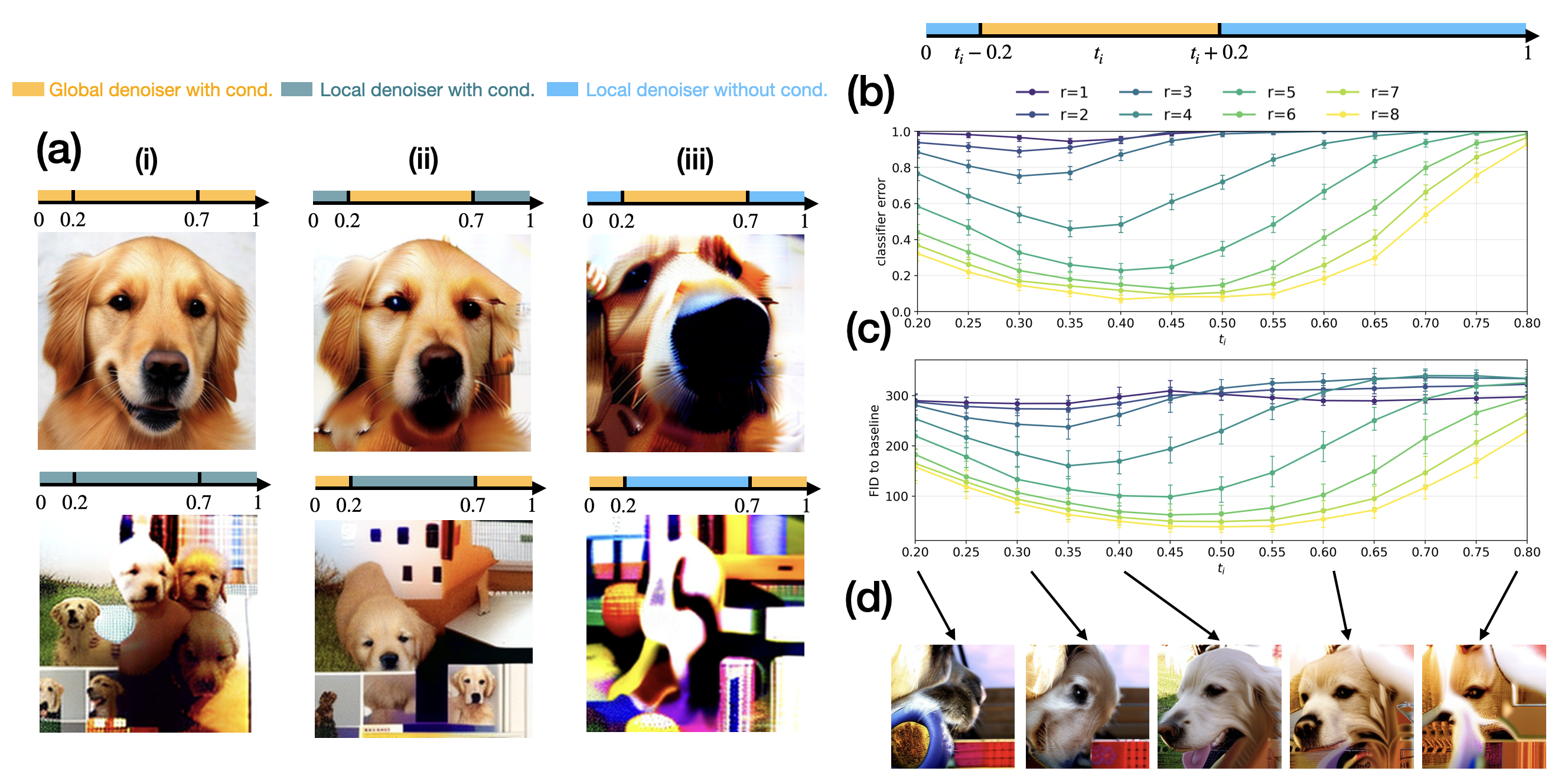}
  \caption{\textbf{Time-windowed local denoising in Facebook DiT-XL}. (a) Golden retriever samples generated with different combinations of local and global denoisers. The local denoiser uses radius $r=3$. (i) Top: global denoiser at all times. Bottom: local denoiser at all times. (ii) Top: global denoiser in $[0.2,0.7]$ and local conditioned denoiser outside the window. Bottom: local conditioned denoiser in $[0.2,0.7]$ and global denoiser outside the window. (iii) Top: global denoiser in $[0.2,0.7]$ and local unconditional denoiser outside the window. Bottom: local unconditional denoiser in $[0.2,0.7]$ and global denoiser outside the window. Right column: samples generated with a global denoiser in $[t_i-0.2,t_i+0.2]$ and a local unconditional denoiser with different $r$ outside the window while scanning $t_i$. (b) Classification error as a function of $t_i$. (c) FID to always-global samples as a function of $t_i$. (d) Visualization of generated samples at different $t_i$ using $r=3$.}
  \label{fig:locality-window-imagenet}
\end{figure}


\section{Conclusion}
Our work provides a unified operational view of two notions of phase transitions in diffusion models. By connecting symmetry breaking induced by conditioning with the onset of nonlocal computation in the score, we show that both phenomena can be detected through the same set of measurable quantities. This yields a concrete diagnostic for identifying when global communication is actually required during sampling, rather than assumed a priori from model design. In particular, the agreement (or mismatch) between instantaneous score-based probes and integrated sampling behavior directly reveals how efficiently a model uses its conditioning signal and attention mechanism over time.


These observations have direct implications for both model design and training. On the architecture side, they motivate designs that explicitly control when and how global communication is enabled—for example, by gating cross-token attention, restricting receptive fields outside the critical window, or structuring conditioning pathways so that symmetry breaking is delayed until it is needed. On the training side, the phase diagram suggests reweighting losses or noise schedules to concentrate learning signal near the critical interval, encouraging the model to align symmetry breaking with the onset of nonlocality. One can also regularize locality (e.g., via masked attention or locality constraints) at early times and gradually relax these constraints during training, effectively shaping the trajectory of the learned score.

More broadly, our framework suggests a principled route to improving diffusion models. It enables systematic comparison across architectures by reducing complex generation behavior to phase diagrams over time, and it supports adaptive sampling schemes that switch between local and global denoisers based on the identified critical window. It also provides concrete guidance for co-designing architectures and training procedures so that useful global computation is both temporally localized and computationally efficient, opening the possibility of achieving comparable or improved generation quality with significantly reduced cost.

\bibliographystyle{plainnat}
\bibliography{reference}

@inproceedings{liu2021swin,
  title     = {Swin Transformer: Hierarchical Vision Transformer using Shifted Windows},
  author    = {Liu, Ze and Lin, Yutong and Cao, Yue and Hu, Han and Wei, Yixuan and Zhang, Zheng and Lin, Stephen and Guo, Baining},
  booktitle = {Proceedings of the IEEE/CVF International Conference on Computer Vision (ICCV)},
  pages     = {10012--10022},
  year      = {2021}
}

@inproceedings{yuan2024ditfastattn,
  title     = {DiTFastAttn: Attention Compression for Diffusion Transformer Models},
  author    = {Yuan, Zhihang and Zhang, Hanling and Lu, Pu and Ning, Xuefei and Zhang, Linfeng and Zhao, Tianchen and Yan, Shengen and Dai, Guohao and Wang, Yu},
  booktitle = {Advances in Neural Information Processing Systems},
  volume    = {37},
  year      = {2024},
  note      = {DOI: 10.52202/079017-0037}
}

@article{jin2025stage,
  title   = {Stage-wise Dynamics of Classifier-Free Guidance in Diffusion Models},
  author  = {Jin, Cheng and Shi, Qitan and Gu, Yuantao},
  journal = {arXiv preprint arXiv:2509.22007},
  year    = {2025}
}

@article{wu2025swin,
  title   = {Swin DiT: Diffusion Transformer using Pseudo Shifted Windows},
  author  = {Wu, Jiafu and Wang, Yabiao and Li, Jian and Peng, Jinlong and Cao, Yun and Wang, Chengjie and Zhang, Jiangning},
  journal = {arXiv preprint arXiv:2505.13219},
  year    = {2025}
}

@inproceedings{kynkaanniemi2024applying,
  title     = {Applying Guidance in a Limited Interval Improves Sample and Distribution Quality in Diffusion Models},
  author    = {Kynk{\"a}{\"a}nniemi, Tuomas and Aittala, Miika and Karras, Tero and Laine, Samuli and Aila, Timo and Lehtinen, Jaakko},
  booktitle = {Advances in Neural Information Processing Systems},
  volume    = {37},
  year      = {2024},
  note      = {DOI: 10.52202/079017-3892}
}

@article{shah2025does,
  title   = {Does Generation Require Memorization? Creative Diffusion Models using Ambient Diffusion},
  author  = {Shah, Kulin and Kalavasis, Alkis and Klivans, Adam R. and Daras, Giannis},
  journal = {arXiv preprint arXiv:2502.21278},
  year    = {2025}
}

@article{sclocchi2025phase,
  title   = {A Phase Transition in Diffusion Models Reveals the Hierarchical Nature of Data},
  author  = {Sclocchi, Antonio and Favero, Alessandro and Wyart, Matthieu},
  journal = {Proceedings of the National Academy of Sciences},
  volume  = {122},
  number  = {1},
  pages   = {e2408799121},
  year    = {2025},
  doi     = {10.1073/pnas.2408799121}
}

@techreport{hassani2023neighborhood,
  title        = {Neighborhood Attention: Dynamic Restriction of Self Attention},
  author       = {Hassani, Ali},
  institution  = {University of Oregon, Department of Computer Science},
  number       = {AREA-202307},
  year         = {2023},
  url          = {https://www.cs.uoregon.edu/Reports/AREA-202307-Hassani.pdf}
}

@inproceedings{hassani2023nat,
  title     = {Neighborhood Attention Transformer},
  author    = {Hassani, Ali and Walton, Steven and Li, Jiachen and Li, Shen and Shi, Humphrey},
  booktitle = {Proceedings of the IEEE/CVF Conference on Computer Vision and Pattern Recognition (CVPR)},
  pages     = {6185--6194},
  year      = {2023},
  doi       = {10.1109/CVPR52729.2023.00599}
}

@misc{hassani2022dilated,
  title         = {Dilated Neighborhood Attention Transformer},
  author        = {Hassani, Ali and Shi, Humphrey},
  year          = {2022},
  eprint        = {2209.15001},
  archivePrefix = {arXiv},
  primaryClass  = {cs.CV},
  url           = {https://arxiv.org/abs/2209.15001}
}

@inproceedings{sclocchi2025probing,
  title     = {Probing the Latent Hierarchical Structure of Data via Diffusion Models},
  author    = {Sclocchi, Antonio and Favero, Alessandro and Levi, Noam Itzhak and Wyart, Matthieu},
  booktitle = {Proceedings of the International Conference on Learning Representations (ICLR)},
  year      = {2025},
  url       = {https://arxiv.org/abs/2410.13770},
  eprint    = {2410.13770},
  archivePrefix = {arXiv},
  primaryClass  = {cs.LG}
}

@InProceedings{sohl-dickstein-diffusion,
  title = 	 {Deep Unsupervised Learning using Nonequilibrium Thermodynamics},
  author = 	 {Sohl-Dickstein, Jascha and Weiss, Eric and Maheswaranathan, Niru and Ganguli, Surya},
  booktitle = 	 {Proceedings of the 32nd International Conference on Machine Learning},
  pages = 	 {2256--2265},
  year = 	 {2015},
  editor = 	 {Bach, Francis and Blei, David},
  volume = 	 {37},
  series = 	 {Proceedings of Machine Learning Research},
  address = 	 {Lille, France},
  month = 	 {07--09 Jul},
  publisher =    {PMLR},
  url = 	 {https://proceedings.mlr.press/v37/sohl-dickstein15.html}
}

@inproceedings{DDPM20,
author = {Ho, Jonathan and Jain, Ajay and Abbeel, Pieter},
title = {Denoising diffusion probabilistic models},
year = {2020},
isbn = {9781713829546},
publisher = {Curran Associates Inc.},
address = {Red Hook, NY, USA},
booktitle = {Proceedings of the 34th International Conference on Neural Information Processing Systems},
articleno = {574},
numpages = {12},
location = {Vancouver, BC, Canada},
series = {NIPS '20}
}

@inproceedings{
scoreSDE2021,
title={Score-Based Generative Modeling through Stochastic Differential Equations},
author={Yang Song and Jascha Sohl-Dickstein and Diederik P Kingma and Abhishek Kumar and Stefano Ermon and Ben Poole},
booktitle={International Conference on Learning Representations},
year={2021},
url={https://openreview.net/forum?id=PxTIG12RRHS}
}

@inproceedings{
SDEedit2022,
title={{SDE}dit: Guided Image Synthesis and Editing with Stochastic Differential Equations},
author={Chenlin Meng and Yutong He and Yang Song and Jiaming Song and Jiajun Wu and Jun-Yan Zhu and Stefano Ermon},
booktitle={International Conference on Learning Representations},
year={2022},
url={https://openreview.net/forum?id=aBsCjcPu_tE}
}

@INPROCEEDINGS{perceptual_imp2022,
  author={Choi, Jooyoung and Lee, Jungbeom and Shin, Chaehun and Kim, Sungwon and Kim, Hyunwoo and Yoon, Sungroh},
  booktitle={2022 IEEE/CVF Conference on Computer Vision and Pattern Recognition (CVPR)}, 
  title={Perception Prioritized Training of Diffusion Models}, 
  year={2022},
  volume={},
  number={},
  pages={11462-11471},
  doi={10.1109/CVPR52688.2022.01118}}

@inproceedings{
raya2023spontaneous,
title={Spontaneous symmetry breaking in generative diffusion models},
author={Gabriel Raya and Luca Ambrogioni},
booktitle={Thirty-seventh Conference on Neural Information Processing Systems},
year={2023},
url={https://openreview.net/forum?id=lxGFGMMSVl}
}

@article{Birolicurieweiss_2023,
   title={Generative diffusion in very large dimensions},
   volume={2023},
   ISSN={1742-5468},
   url={http://dx.doi.org/10.1088/1742-5468/acf8ba},
   DOI={10.1088/1742-5468/acf8ba},
   number={9},
   journal={Journal of Statistical Mechanics: Theory and Experiment},
   publisher={IOP Publishing},
   author={Biroli, Giulio and Mézard, Marc},
   year={2023},
   month=sep, pages={093402} }

@Article{Ambrogioni_stochastic_2025,
AUTHOR = {Ambrogioni, Luca},
TITLE = {The Statistical Thermodynamics of Generative Diffusion Models: Phase Transitions, Symmetry Breaking, and Critical Instability},
JOURNAL = {Entropy},
VOLUME = {27},
YEAR = {2025},
NUMBER = {3},
ARTICLE-NUMBER = {291},
URL = {https://www.mdpi.com/1099-4300/27/3/291},
PubMedID = {40149215},
ISSN = {1099-4300},
DOI = {10.3390/e27030291}
}

@article{dynamical_regimes,
	title = {Dynamical regimes of diffusion models},
	volume = {15},
	issn = {2041-1723},
	url = {https://doi.org/10.1038/s41467-024-54281-3},
	doi = {10.1038/s41467-024-54281-3},
	number = {1},
	journal = {Nature Communications},
	author = {Biroli, Giulio and Bonnaire, Tony and de Bortoli, Valentin and Mézard, Marc},
	month = nov,
	year = {2024},
	pages = {9957},
}

@inproceedings{critical_windows,
author = {Li, Marvin and Chen, Sitan},
title = {Critical windows: non-asymptotic theory for feature emergence in diffusion models},
year = {2024},
publisher = {JMLR.org},
booktitle = {Proceedings of the 41st International Conference on Machine Learning},
articleno = {1097},
numpages = {25},
location = {Vienna, Austria},
series = {ICML'24}
}

@article{
hierarchical_diffusion,
author = {Antonio Sclocchi  and Alessandro Favero  and Matthieu Wyart },
title = {A phase transition in diffusion models reveals the hierarchical nature of data},
journal = {Proceedings of the National Academy of Sciences},
volume = {122},
number = {1},
pages = {e2408799121},
year = {2025},
doi = {10.1073/pnas.2408799121},
URL = {https://www.pnas.org/doi/abs/10.1073/pnas.2408799121},
eprint = {https://www.pnas.org/doi/pdf/10.1073/pnas.2408799121}}

@misc{ventura_distortion_diff,
      title={Emergence of Distortions in High-Dimensional Guided Diffusion Models}, 
      author={Enrico Ventura and Beatrice Achilli and Luca Ambrogioni and Carlo Lucibello},
      year={2026},
      eprint={2602.00716},
      archivePrefix={arXiv},
      primaryClass={stat.ML},
      url={https://arxiv.org/abs/2602.00716}, 
}

@article{
yaguchi_geometry2025,
title={The Geometry of Phase Transitions in Diffusion Models: Tubular Neighbourhoods and Singularities},
author={Manato Yaguchi and Kotaro Sakamoto and Ryosuke Sakamoto and Masato Tanabe and Masatomo Akagawa and Yusuke Hayashi and Masahiro Suzuki and Yutaka Matsuo},
journal={Transactions on Machine Learning Research},
issn={2835-8856},
year={2025},
url={https://openreview.net/forum?id=ahVFKFLYk2},
note={Featured Certification}
}

@inproceedings{
pham2024memorization,
title={Memorization to Generalization: The Emergence of Diffusion Models from Associative Memory},
author={Bao Pham and Gabriel Raya and Matteo Negri and Mohammed J Zaki and Luca Ambrogioni and Dmitry Krotov},
booktitle={NeurIPS 2024 Workshop on Scientific Methods for Understanding Deep Learning},
year={2024},
url={https://openreview.net/forum?id=zVMMaVy2BY}
}

@inproceedings{
li2025blink,
title={Blink of an eye: a simple theory for feature localization in generative models},
author={Marvin Li and Aayush Karan and Sitan Chen},
booktitle={Forty-second International Conference on Machine Learning},
year={2025},
url={https://openreview.net/forum?id=QvqnPVGWAN}
}

@misc{handke2025measuringsemanticinformationproduction,
      title={Measuring Semantic Information Production in Generative Diffusion Models}, 
      author={Florian Handke and Félix Koulischer and Gabriel Raya and Luca Ambrogioni},
      year={2025},
      eprint={2506.10433},
      archivePrefix={arXiv},
      primaryClass={stat.ML},
      url={https://arxiv.org/abs/2506.10433}, 
}

@Article{info_dynamics_diffusion,
AUTHOR = {Stančević, Dejan and Ambrogioni, Luca},
TITLE = {The Information Dynamics of Generative Diffusion},
JOURNAL = {Entropy},
VOLUME = {28},
YEAR = {2026},
NUMBER = {2},
ARTICLE-NUMBER = {195},
URL = {https://www.mdpi.com/1099-4300/28/2/195},
PubMedID = {41751698},
ISSN = {1099-4300},
DOI = {10.3390/e28020195}
}

@inproceedings{
ramachandran2026crossfluctuation,
title={Cross-fluctuation phase transitions reveal sampling dynamics in diffusion models},
author={Sai Niranjan Ramachandran and Manish Krishan Lal and Suvrit Sra},
booktitle={The Thirty-ninth Annual Conference on Neural Information Processing Systems},
year={2026},
url={https://openreview.net/forum?id=b4X6cz1F9l}
}

@misc{lu2026steeringdynamicalregimesdiffusion,
      title={Steering Dynamical Regimes of Diffusion Models by Breaking Detailed Balance}, 
      author={Haiqi Lu and Ying Tang},
      year={2026},
      eprint={2602.15914},
      archivePrefix={arXiv},
      primaryClass={cond-mat.stat-mech},
      url={https://arxiv.org/abs/2602.15914}, 
}

@misc{handke2026entropicsignatureclassspeciation,
      title={The Entropic Signature of Class Speciation in Diffusion Models}, 
      author={Florian Handke and Dejan Stančević and Felix Koulischer and Thomas Demeester and Luca Ambrogioni},
      year={2026},
      eprint={2602.09651},
      archivePrefix={arXiv},
      primaryClass={stat.ML},
      url={https://arxiv.org/abs/2602.09651}, 
}

@misc{ambrogioni2026outofequilibriumphasetransitionsseed,
      title={How Out-of-Equilibrium Phase Transitions can Seed Pattern Formation in Trained Diffusion Models}, 
      author={Luca Ambrogioni},
      year={2026},
      eprint={2603.20092},
      archivePrefix={arXiv},
      primaryClass={cs.LG},
      url={https://arxiv.org/abs/2603.20092}, 
}

@misc{takahashi2026dynamicalregimesdiscretediffusion,
      title={Dynamical Regimes of Discrete Diffusion Models}, 
      author={Tomoei Takahashi and Takashi Takahashi and Yoshiyuki Kabashima},
      year={2026},
      eprint={2604.10961},
      archivePrefix={arXiv},
      primaryClass={cond-mat.stat-mech},
      url={https://arxiv.org/abs/2604.10961}, 
}

@article{sang_mixed_phase,
  title = {Stability of Mixed-State Quantum Phases via Finite Markov Length},
  author = {Sang, Shengqi and Hsieh, Timothy H.},
  journal = {Phys. Rev. Lett.},
  volume = {134},
  issue = {7},
  pages = {070403},
  numpages = {7},
  year = {2025},
  month = {Feb},
  publisher = {American Physical Society},
  doi = {10.1103/PhysRevLett.134.070403},
  url = {https://link.aps.org/doi/10.1103/PhysRevLett.134.070403}
}

@misc{zhang2025conditionalmutualinformationinformationtheoretic,
      title={Conditional Mutual Information and Information-Theoretic Phases of Decohered Gibbs States}, 
      author={Yifan Zhang and Sarang Gopalakrishnan},
      year={2025},
      eprint={2502.13210},
      archivePrefix={arXiv},
      primaryClass={quant-ph},
      url={https://arxiv.org/abs/2502.13210}, 
}

@misc{zhang2025stabilitymixedstatephasesweak,
      title={Stability of mixed-state phases under weak decoherence}, 
      author={Yifan F. Zhang and Sarang Gopalakrishnan},
      year={2025},
      eprint={2511.01976},
      archivePrefix={arXiv},
      primaryClass={quant-ph},
      url={https://arxiv.org/abs/2511.01976}, 
}

@article{cmi_decay_mixed,
  title = {Universal Decay of Mutual Information and Conditional Mutual Information in Gapped Pure- and Mixed-State Quantum Matter},
  author = {Yi, Jinmin and Li, Kangle and Liu, Chuan and Li, Zixuan and Zou, Liujun},
  journal = {Phys. Rev. Lett.},
  volume = {136},
  issue = {11},
  pages = {116604},
  numpages = {8},
  year = {2026},
  month = {Mar},
  publisher = {American Physical Society},
  doi = {10.1103/mqp8-y1m7},
  url = {https://link.aps.org/doi/10.1103/mqp8-y1m7}
}

@misc{hu2025localdiffusionmodelsphases,
      title={Local Diffusion Models and Phases of Data Distributions}, 
      author={Fangjun Hu and Guangkuo Liu and Yifan F. Zhang and Xun Gao},
      year={2025},
      eprint={2508.06614},
      archivePrefix={arXiv},
      primaryClass={cs.LG},
      url={https://arxiv.org/abs/2508.06614}, 
}

@inproceedings{perez2018film,
  title={Film: Visual reasoning with a general conditioning layer},
  author={Perez, Ethan and Strub, Florian and De Vries, Harm and Dumoulin, Vincent and Courville, Aaron},
  booktitle={Proceedings of the AAAI conference on artificial intelligence},
  volume={32},
  year={2018}
}

@inproceedings{peebles2023scalable,
  title={Scalable diffusion models with transformers},
  author={Peebles, William and Xie, Saining},
  booktitle={Proceedings of the IEEE/CVF international conference on computer vision},
  pages={4195--4205},
  year={2023}
}

@article{han2025entropic,
  title={Entropic order},
  author={Han, Yiqiu and Huang, Xiaoyang and Komargodski, Zohar and Lucas, Andrew and Popov, Fedor K},
  journal={Nature Communications},
  year={2025},
  publisher={Nature Publishing Group UK London}
}

@article{ho2022classifier,
  title={Classifier-free diffusion guidance},
  author={Ho, Jonathan and Salimans, Tim},
  journal={arXiv preprint arXiv:2207.12598},
  year={2022}
}

@article{Mermin-Wagner,
  title = {Absence of Ferromagnetism or Antiferromagnetism in One- or Two-Dimensional Isotropic Heisenberg Models},
  author = {Mermin, N. D. and Wagner, H.},
  journal = {Phys. Rev. Lett.},
  volume = {17},
  issue = {22},
  pages = {1133--1136},
  numpages = {0},
  year = {1966},
  month = {Nov},
  publisher = {American Physical Society},
  doi = {10.1103/PhysRevLett.17.1133},
  url = {https://link.aps.org/doi/10.1103/PhysRevLett.17.1133}
}

@article{kamb2024analytic,
  title={An analytic theory of creativity in convolutional diffusion models},
  author={Kamb, Mason and Ganguli, Surya},
  journal={arXiv preprint arXiv:2412.20292},
  year={2024}
}

@article{lukoianov2025locality,
  title={Locality in image diffusion models emerges from data statistics},
  author={Lukoianov, Artem and Yuan, Chenyang and Solomon, Justin and Sitzmann, Vincent},
  journal={arXiv preprint arXiv:2509.09672},
  year={2025}
}

@article{niedoba2024towards,
  title={Towards a mechanistic explanation of diffusion model generalization},
  author={Niedoba, Matthew and Zwartsenberg, Berend and Murphy, Kevin and Wood, Frank},
  journal={arXiv preprint arXiv:2411.19339},
  year={2024}
}

@misc{ddim2022,
      title={Denoising Diffusion Implicit Models}, 
      author={Jiaming Song and Chenlin Meng and Stefano Ermon},
      year={2022},
      eprint={2010.02502},
      archivePrefix={arXiv},
      primaryClass={cs.LG},
      url={https://arxiv.org/abs/2010.02502}, 
}

@misc{esser2024scalingrectifiedflowtransformers,
      title={Scaling Rectified Flow Transformers for High-Resolution Image Synthesis}, 
      author={Patrick Esser and Sumith Kulal and Andreas Blattmann and Rahim Entezari and Jonas Müller and Harry Saini and Yam Levi and Dominik Lorenz and Axel Sauer and Frederic Boesel and Dustin Podell and Tim Dockhorn and Zion English and Kyle Lacey and Alex Goodwin and Yannik Marek and Robin Rombach},
      year={2024},
      eprint={2403.03206},
      archivePrefix={arXiv},
      primaryClass={cs.CV},
      url={https://arxiv.org/abs/2403.03206}, 
}

@article{lu2022dpm,
  title={Dpm-solver: A fast ode solver for diffusion probabilistic model sampling in around 10 steps},
  author={Lu, Cheng and Zhou, Yuhao and Bao, Fan and Chen, Jianfei and Li, Chongxuan and Zhu, Jun},
  journal={Advances in neural information processing systems},
  volume={35},
  pages={5775--5787},
  year={2022}
}

@article{lu2025dpm,
  title={Dpm-solver++: Fast solver for guided sampling of diffusion probabilistic models},
  author={Lu, Cheng and Zhou, Yuhao and Bao, Fan and Chen, Jianfei and Li, Chongxuan and Zhu, Jun},
  journal={Machine Intelligence Research},
  volume={22},
  number={4},
  pages={730--751},
  year={2025},
  publisher={Springer}
}

@inproceedings{he2016deep,
  title={Deep residual learning for image recognition},
  author={He, Kaiming and Zhang, Xiangyu and Ren, Shaoqing and Sun, Jian},
  booktitle={Proceedings of the IEEE conference on computer vision and pattern recognition},
  pages={770--778},
  year={2016}
}

@misc{resnet50_hf,
title = {ResNet-50 v1.5},
author = {{Hugging Face} and {Microsoft}},
howpublished = {\url{https://huggingface.co/microsoft/resnet-50}},
year = {2025}
}

@article{hu_2026_learning,
  url = {https://arxiv.org/abs/2604.01197},
  author = {Hu,  Fangjun and Kokail,  Christian and Kornjača,  Milan and Lopes,  Pedro L. S. and Gong,  Weiyuan and Wang,  Sheng-Tao and Gao,  Xun and Ostermann,  Stefan},
  title = {Learning and Generating Mixed States Prepared by Shallow Channel Circuits},
  year = {2026},
  journal={arXiv:2604.01197 [quant-ph]},
}

@article{kumar_2026_unlearnable,
  url = {https://arxiv.org/abs/2602.11262},
  author = {Kumar,  Tarun Advaith and Zou,  Yijian and Negari,  Amir-Reza and Melko,  Roger G. and Hsieh,  Timothy H.},
  title = {Unlearnable phases of matter},
  journal={arXiv:2602.11262 [cond-mat.dis-nn]},
  year={2026}
}

@inproceedings{deng2009imagenet,
  title={Imagenet: A large-scale hierarchical image database},
  author={Deng, Jia and Dong, Wei and Socher, Richard and Li, Li-Jia and Li, Kai and Fei-Fei, Li},
  booktitle={2009 IEEE conference on computer vision and pattern recognition},
  pages={248--255},
  year={2009},
  organization={Ieee}
}

@article{heusel2017gans,
  title={Gans trained by a two time-scale update rule converge to a local nash equilibrium},
  author={Heusel, Martin and Ramsauer, Hubert and Unterthiner, Thomas and Nessler, Bernhard and Hochreiter, Sepp},
  journal={Advances in neural information processing systems},
  volume={30},
  year={2017}
}

@inproceedings{szegedy2016rethinking,
  title={Rethinking the inception architecture for computer vision},
  author={Szegedy, Christian and Vanhoucke, Vincent and Ioffe, Sergey and Shlens, Jon and Wojna, Zbigniew},
  booktitle={Proceedings of the IEEE conference on computer vision and pattern recognition},
  pages={2818--2826},
  year={2016}
}


\appendix

\section{Additional Data}

\begin{figure}
  \centering
  \includegraphics[width=\linewidth]{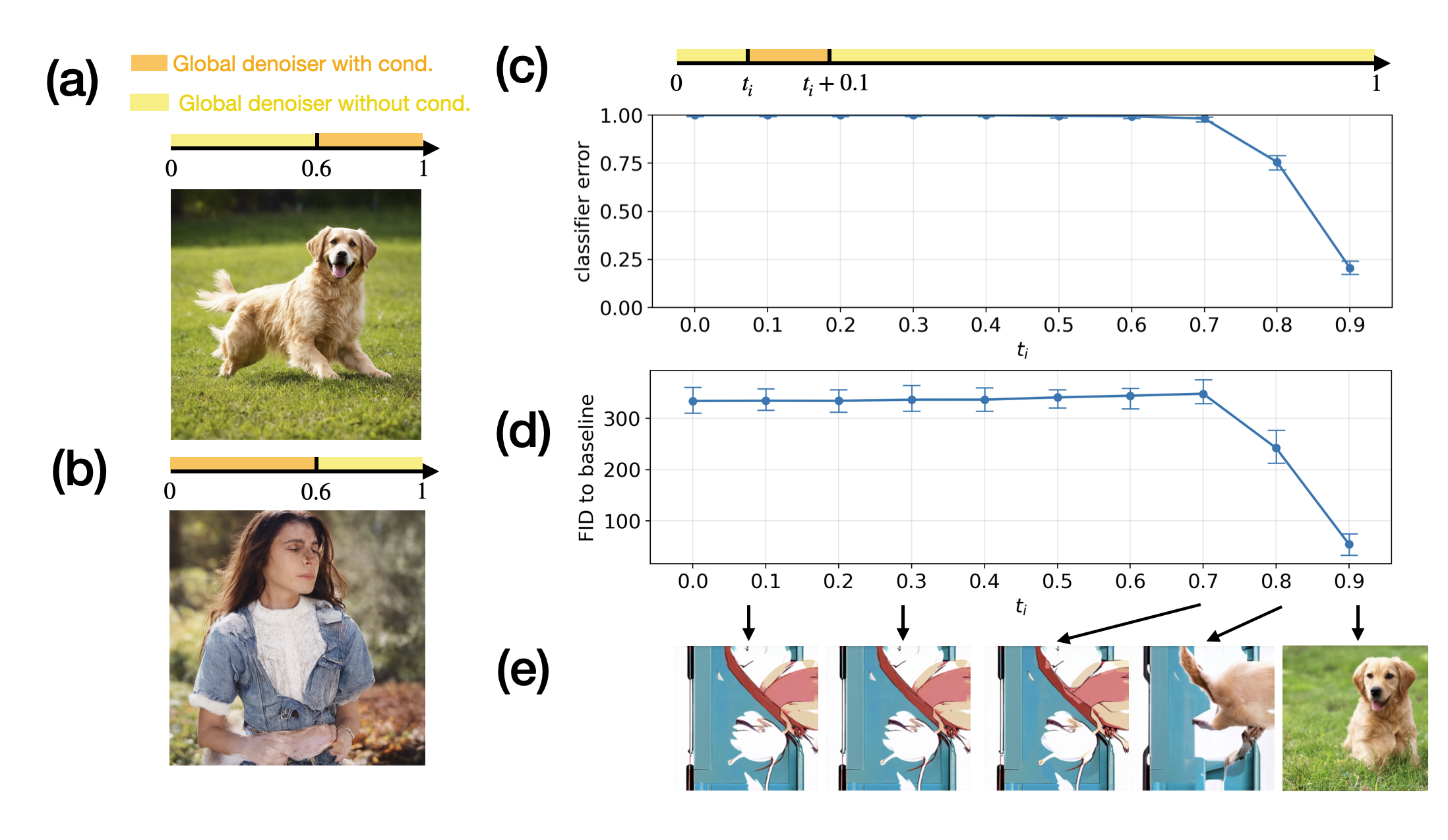}
  \caption{\textbf{Windowed conditioning in SD3 medium}. Left column: (a) Golden retriever sample generated with conditioning applied only in the time window $[0.6,1.0]$ and no conditioning outside the window. (b) Sample generated with conditioning removed in $[0.6,1.0]$ and applied outside the window. Right column: samples generated with conditioning applied only in short windows $[t_i,t_i+0.1]$ while scanning $t_i$. (c) Classification error as a function of $t_i$. (d) FID to always-conditioned samples as a function of $t_i$. (e) Visualization of generated samples at different $t_i$.}
  \label{fig:condition-window-sd3}
\end{figure}

\begin{figure}
  \centering
  \includegraphics[width=\linewidth]{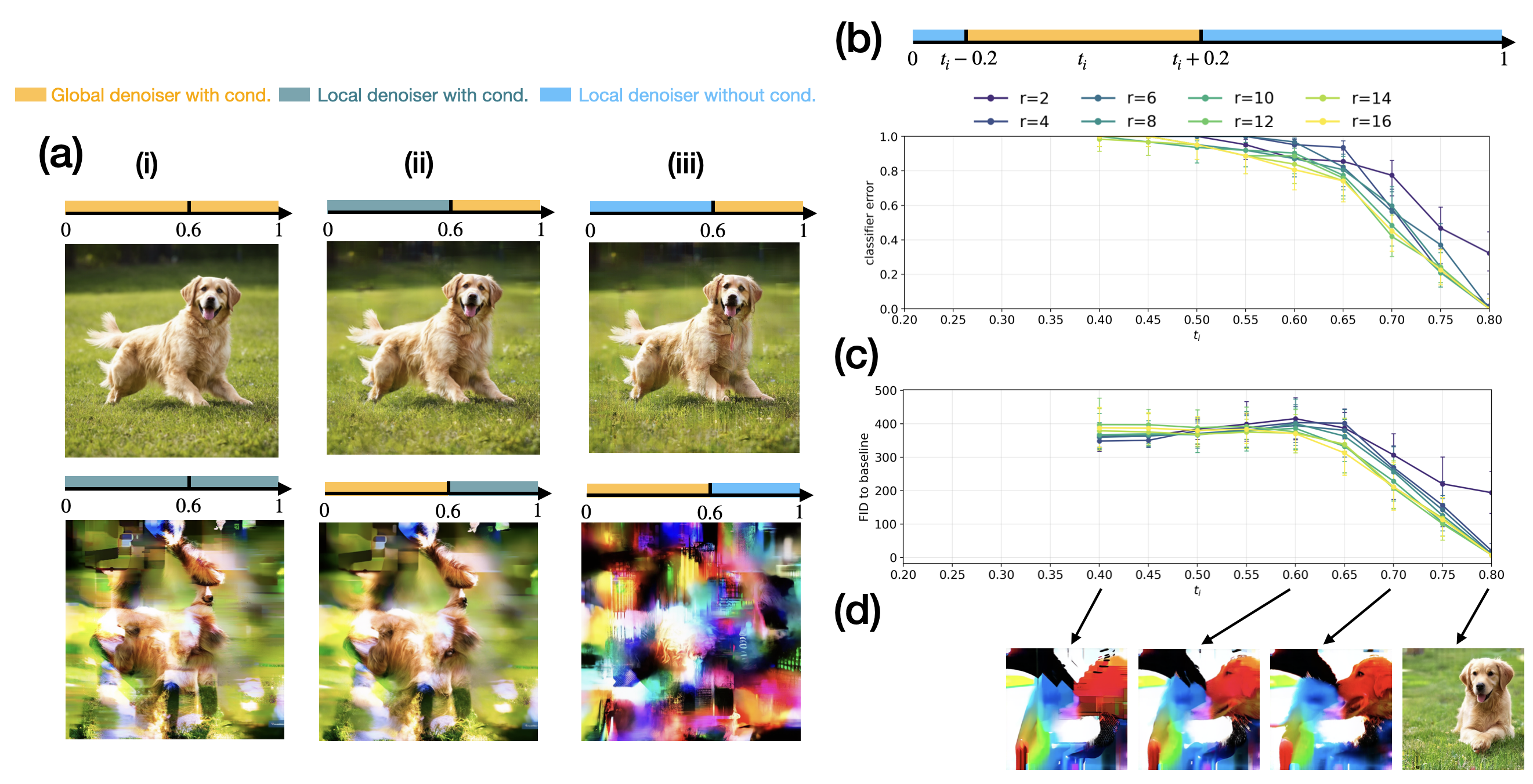}
  \caption{\textbf{Windowed local denoising in SD3 medium}. (a) Golden retriever samples generated with different combinations of local and global denoisers. The local denoiser uses radius $r=6$. (i) Top: global denoiser at all times. Bottom: local denoiser at all times. (ii) Top: global denoiser in $[0.6,1.0]$ and local conditioned denoiser outside the window. Bottom: local conditioned denoiser in $[0.6,1.0]$ and global denoiser outside the window. (iii) Top: global denoiser in $[0.6,1.0]$ and local unconditional denoiser outside the window. Bottom: local unconditional denoiser in $[0.6,1.0]$ and global denoiser with different $r$ outside the window. Right column: samples generated with a global denoiser in $[t_i-0.2,t_i+0.2]$ and a local unconditional denoiser outside the window while scanning $t_i$. (b) Classification error as a function of $t_i$. (c) FID to always-global samples as a function of $t_i$. (d) Visualization of generated samples at different $t_i$ using $r=6$.}
  \label{fig:locality-window-sd3}
\end{figure}

Figure~\ref{fig:condition-window-sd3} and Figure~\ref{fig:locality-window-sd3} provide the SD3 medium versions of the two windowed sampling experiments discussed in the main text.

\begin{figure}
  \centering
  \includegraphics[width=\linewidth]{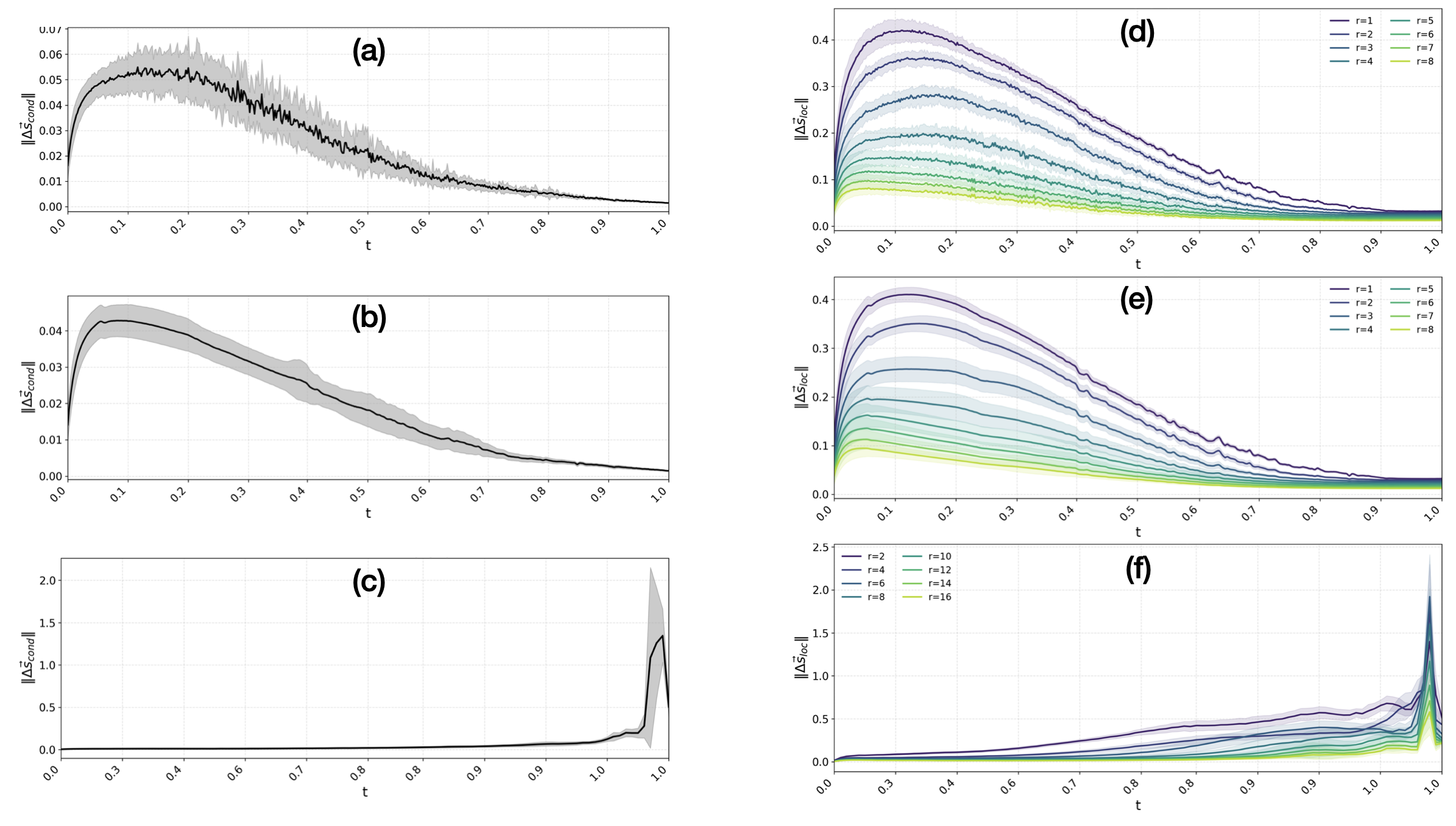}
  \caption{\textbf{Fluctution of score gap}. Plot of score gaps with colored area marking the standard deviation across different trajectories. (a) Conditioning gap and (b) locality for Facenook DiT along the training trajectory. (c) Conditioning gap and (d) locality gap for Facenook DiT along the conditional sampling trajectory. (e) Conditioning gap and (f) locality gap for SD3 medium along the conditional sampling trajectory. }
  \label{fig:score_gap_fluctuation}
\end{figure}

One may ask if the score gap plots are typical behaviors for different trajectories. Figure~\ref{fig:score_gap_fluctuation} plots the standard deviation of different score gaps due to the randomness in the trajectory. We observe a small fluctuation across all score gaps, indicating that the observed behavior is typical.

\begin{figure}
  \centering
  \includegraphics[width=\linewidth]{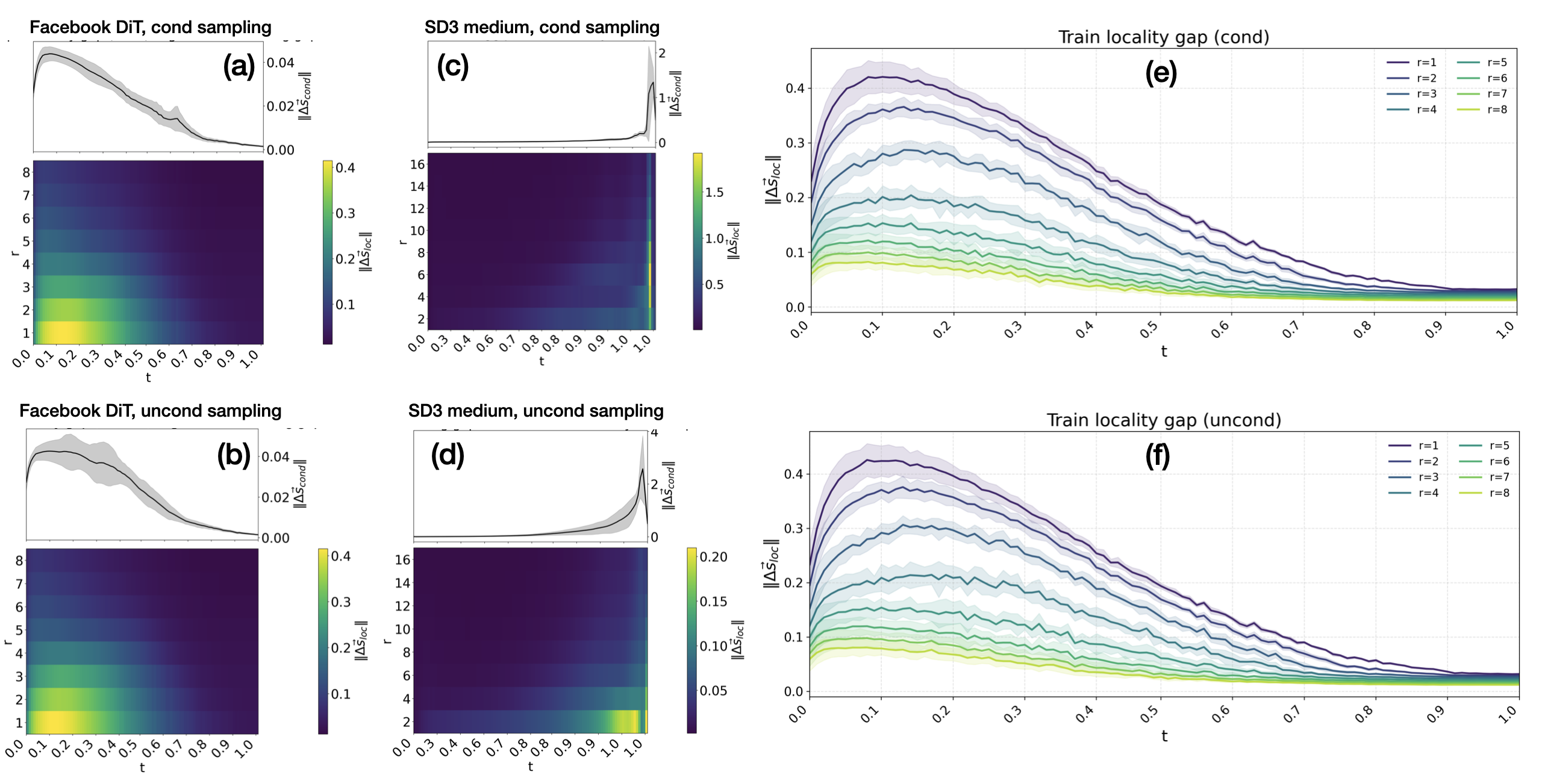}
  \caption{\textbf{Score gap along unconditional trajectories}. (a) Conditioning gap (top) and locality gap defined using conditional score function, for Facebook DiT along the conditional sampling trajectory. (b) Conditioning gap (top) and locality gap defined using unconditional score function, for Facebook DiT along the unconditional sampling trajectory. (c) Conditioning gap (top) and locality gap defined using conditional score function, for SD3 medium along the conditional sampling trajectory. (d) Conditioning gap (top) and locality gap defined using unconditional score function, for SD3 medium along the unconditional sampling trajectory. (e,f) Locality gap for Facebook DiT along the training trajectory, for (e) conditional and (f) unconditional score functions. (e,f) are visually indistinguishable because their differences are around 0.01. All colored areas mark the standard deviation across different trajectories.}
  \label{fig:score_gap_uncond}
\end{figure}

We also provide additional data comparing the score gaps along conditional and unconditional sampling trajectories in Figure~\ref{fig:score_gap_uncond}(a-d). For conditional sampling trajectories we report the conditional locality gap since it is relevant to this type of trajectory. Similarly, for unconditional sampling trajectories we report the unconditional locality gap. We observe the same behavior for both types of trajectories. We also compare the conditional and unconditional locality gap for Facebook DiT along the training trajectory. We observe a very small difference.

\section{Detailed Review of Earlier Work and Theoretical Background}
We give a more detailed review of earlier work and theoretical background in this section. 
\subsection{Review of earlier work}
Using a diffusion process for generative modeling was proposed by
~\cite{sohl-dickstein-diffusion} and their modern formulations as DDPM and score-SDE were set in 
~\cite{DDPM20} and
~\cite{scoreSDE2021}, respectively. Early empirical work, such as~\cite{perceptual_imp2022} or~\cite{SDEedit2022}, already noted that information injection during denoising is not uniform in time,
and these observations rapidly inspired a detailed study into dynamics of denoising.

Its origin in non-equilibrium physics invited a study of generative diffusion via a statistical physics lens.~\cite{raya2023spontaneous} found hints of phase transitions in diffusion models, and further work 
by~\cite{Birolicurieweiss_2023} and~\cite{Ambrogioni_stochastic_2025} 
analyze symmetry breaking (origin of phase transitions) toy models for data distributions. 
Existence of a phase transition in generation motivates understanding the full denoising process, which was analyzed asymptotically by~\cite{dynamical_regimes},
and non-asymptotically by~\cite{critical_windows}
for when features emerge during generation. 
Other works by ~\cite{hierarchical_diffusion},~\cite{ventura_distortion_diff},~\cite{yaguchi_geometry2025} studied details of the generation process, such as how various regimes and their time-lengths may be determined the structure of the data itself.

Recent work expands this phase-transition view toward memory, information diagnostics, and control.~\cite{pham2024memorization} study the transition from memorization to generalization through associative-memory spurious states, 
and~\cite{li2025blink} 
develop a broader feature-localization theory for generative models. 
\cite{handke2025measuringsemanticinformationproduction} 
measure class-semantic information production along diffusion trajectories, and ~\cite{info_dynamics_diffusion}
connect entropy production, score divergence, and branching dynamics, 
and \cite{ramachandran2026crossfluctuation} use cross-fluctuations to detect sampling transitions and improve downstream generation and classification.
~\cite{lu2026steeringdynamicalregimesdiffusion} study how nonreversible perturbations can steer dynamical regimes by breaking detailed balance. 
Very recent extensions study entropic signatures of class speciation~\cite{handke2026entropicsignatureclassspeciation},
out-of-equilibrium pattern formation in trained diffusion models~\cite{ambrogioni2026outofequilibriumphasetransitionsseed},
and analogous regimes in discrete diffusion models~\cite{takahashi2026dynamicalregimesdiscretediffusion}. 
Our symmetry breaking probe belongs to this lineage, but instead of inferring semantic commitment from generated samples, we directly measure the classifier-free guidance (CFG) score correction $s_\theta(x_t,t\mid y)-s_\theta(x_t,t)$ as a model-internal field.

A separate line of work characterizes phases by conditional independence and local recoverability. Given two distributions,~\cite{sang_mixed_phase} proposed the decay length-scale of conditional mutual information (CMI) (called the `Markov length') as a diagnostic of mixed-state phase transitions, which diverges if the two are in different phases and hence not locally recoverable. 
~\cite{zhang2025stabilitymixedstatephasesweak} proved that under weak local decoherence, the mixed-state phases are stable, and~\cite{cmi_decay_mixed} proved that finite Markov length is a generic feature of mixed-state phases. Further work by~\cite{zhang2025conditionalmutualinformationinformationtheoretic} 
proved that the Markov length is finite during diffusion for high-temperature Gibbs states while noting that at low temperature, it may diverge. 

\cite{hu2025localdiffusionmodelsphases} brought this Markov-length perspective directly to diffusion models by defining phases of data distributions analogously to mixed state phases: between two phases, local denoisers fail near a narrow transition window where Markov length diverges, while it remains finite away from this window, which was shown explicitly by calculating the CMI during image generation. 
Prior work by \cite{hu_2026_learning} has further shown that if a distribution admits a diffusion noise path along which the CMI remains short-ranged, then one can construct a generative procedure that samples from this distribution using only local operations. Conversely, distributions with long-ranged CMI are known to be hard to learn \cite{kumar_2026_unlearnable}.
Taken together, these results suggest that the Markov length provides a quantitative indicator of the difficulty of learning and generating a distribution.
Our locality-gap heatmaps are an operational analogue of their CMI heatmaps: instead of estimating an information-theoretic decoder, we replace the global denoiser by a local-attention denoiser and directly measure the resulting score and sample degradation. The local-attention intervention itself is related to Swin Transformer by~\cite{liu2021swin}, Neighborhood Attention Transformer by~\cite{hassani2023nat}, neighborhood attention and dilated neighborhood attention by~\cite{hassani2023neighborhood,hassani2022dilated}, DiTFastAttn by~\cite{yuan2024ditfastattn}, and Swin DiT by~\cite{wu2025swin}, but our use is diagnostic rather than architectural.

Our contribution is to connect the two phase-transition pictures experimentally. Prior diffusion work asks when semantic decisions, speciation, or collapse occur; prior Markov-length work asks when local neighborhoods suffice for recovery or denoising. We ask whether these two criticalities align in real diffusion transformers by comparing conditioning gaps with locality gaps at the score level and with error-correction and windowed-sampling behavior at the sample level. This also connects to limited-interval guidance by~\cite{kynkaanniemi2024applying}, stage-wise CFG dynamics by~\cite{jin2025stage}, ambient-diffusion studies of memorization by~\cite{shah2025does}, and latent-hierarchy probing by ~\cite{sclocchi2025probing}.

Understanding locality bias has also been a recurrent topic in literature. \cite{kamb2024analytic} shows that convolutional diffusion model exploits locality bias explicitly which leads to generalization / creativity. \cite{niedoba2024towards} similarly shows the emergence of locality bias in convolutional model and DiT and designs mechanistically interpretable local denoisers to approximate global denoisers. \cite{lukoianov2025locality} shows that even for architectures like transformer that are ignorant of locality, they can learn locality structure through data statistics. Intriguingly, while their numerical technique is drastically different, they observe a time window in which model exhibits long-range correlations. This already alludes to the existence of the nonlocality phase transition. Complement to these work, our main contribution is to understand \emph{when} locality bias can be exploited, and how it correlates with symmetry breaking and semantic forcing.

\subsection{Local computation of the score under a Markov property}
We provide a more detailed theoretical background about local denoising under approximate Markov property.
We encode locality via a rectangular tripartition into the image vector $X=(x_A,x_B,x_C)$ as shown in Figure~\ref{fig:setup}: $A$ is the center patch in the black box, $B$ is the surrounding buffer in gray box, and $C$ is the rest of the image that we faint out. We would like to understand when local generation of $A$ is possible, i.e., when the score on region $A$ can be determined only by the information in $B$, without needing to access $C$. This is the case if the distribution $p(x_A,x_B,x_C)$ forms a Markov chain
\begin{equation}
 x_A \;\text{--}\; x_B \;\text{--}\; x_C,
\quad\text{i.e.}\quad
p(x_A\mid x_B,x_C)=p(x_A\mid x_B),
\end{equation}
which we call the \emph{Markov property}. Then the score on region $A$ depends only on $(x_A,x_B)$:
\begin{align}
\nabla_{x_A}\log p(x_A,x_B,x_C)
&= \nabla_{x_A}\log\Bigl(p(x_A\mid x_B)\,p(x_B,x_C)\Bigr)\notag \\
&= \nabla_{x_A}\log p(x_A\mid x_B).
\end{align}
This motivates a local score approximation to $s_A(x_A,x_B, x_C)$, where
\begin{equation}
 s^{\mathrm{local}}_A(x_A,x_B) = \nabla_{x_A}\log p(x_A\mid x_B),
\end{equation}
which is exact whenever the Markov property holds. 
We expect the Markov property to hold (at least approximately) in real images. For example, to generate the nose of a dog, the model may only need to look at a local neighborhood containing the snout and eyes, and not the rest of the image.

In practice, the Markov property is only approximate. A standard way to quantify this is the conditional mutual information (CMI)
\begin{equation}
 I(A:C\mid B) = \mathbb{E}_{p(x_A,x_B,x_C)}\!\left[\log\frac{p(x_A,x_C\mid x_B)}{p(x_A\mid x_B)p(x_C\mid x_B)}\right].
\end{equation}
We define the Markov length $\xi$ as the length-scale of the empirical decay of CMI with separation: for a neighborhood $B$ of radius $r$ around $A$ and its complement $C$, we posit
\begin{equation}
 I\bigl(A:C\mid B\bigr) \approx I_0\,\exp\!\left(-d_{AC}/\xi\right),
\end{equation}
where $d_{AC}$ is a distance between regions $A$ and $C$ (typically $d_{AC}=r$ when $B_r$ is an $r$-neighborhood). Intuitively, $\xi$ is the length scale beyond which information outside the buffer $B$ is negligible for predicting $A$. This was formalised by \cite{hu2025localdiffusionmodelsphases} as Theorem 1 therein:

\begin{proposition}[Local denoising from finite Markov length]
Consider the $L_x \times L_y$ image space $\smash{X \in \mathbb R^{L_x \cdot L_y}}$. Fix a diffusion time $t$ and let $p_t (X)$ denote the noisy image distribution. Suppose the Markov length $\xi(t) \le \xi_{\max}$ for all $t \le t_c$. Then a local denoiser with radius $O(\xi_{\max} \log(L_x L_y /\epsilon))$ that can produce $p_0(X)$ given $p_t(X)$ with $t \le t_c$, up to total variation distance of $\epsilon$.
\end{proposition}

In general, even if the data distribution satisfies the Markov property exactly, the noised distribution $p_t$ may not, and the Markov length $\xi(t)$ may vary, perhaps even diverge, as a function of $t$~\cite{hu2025localdiffusionmodelsphases}. The divergence of $\xi(t)$ at some critical time $t_c$ is understood as a nonlocality phase transition.

Several theory work have characterized how $\xi(t)$ behaves in discrete spin systems, in both classical and quantum settings. \cite{zhang2025conditionalmutualinformationinformationtheoretic} shows that for high-temperature gibbs state, $\xi(t)$ remains finite for all $t$ up to one. This indicates that if a state is in the high-temperature trival thermodynamic phase, then nonlocality phase transition cannot happen during diffusion. At low temperature, $\xi(t)$ can diverge at some $t_c$ less than one. In fact, one can prove that it has to if the state exhibit long-range two-point correlations, through a simple light-cone argument. This indicates that nonlocality phase transition can be inevitable in sampling low-temperature distributions. Nevertheless, \cite{zhang2025stabilitymixedstatephasesweak} proves that $t_c$ has to be bounded away from zero by a constant amount, indiciating that the low-temperature phase is \emph{stable}. Operationally, this implies that a local denoiser should always exist at sufficiently low noise rate.

So far, all existing rigorous results only applies to discrete spin systems and extending the results to continuous systems remains an open challenge. We expect many of the proof technique to break down due to physical reasons, such as the existence of Goldstone mode which challenges stability~\cite{Mermin-Wagner}
, and the existence of entropic ordering at high temperature, which challenges the lack of phase transitions~\cite{han2025entropic}. These examples are fine-tuned so we do not expect them to appear in realistic datasets. However, any rigorous results and proof technique would have to be able to rule out these adversarial examples. 

\section{Details on Experimental Methodology}

\label{sec:experimental-methodology}

We provide details about all experiments in this section. All experiment are run on at most a single 80G A100 GPU. Most jobs are run a 1/7 MIG partition of a 80G A100 GPU. All job times are less than a day.

\subsection{Local attention implementations}

All local-attention experiments use the same pretrained backbone as the corresponding global baseline and modify only the attention pattern. We replace the attention layer with a radius-$r$ Chebyshev window on the 2D token grid. Queries, keys, and values are projected with the original layer weights; for each image token, the module gathers only keys and values whose grid coordinates satisfy $\max(|\Delta x|, |\Delta y|)\le r$, applies softmax within that window, and reuses the original output projection. For MMDiT joint attention: image-to-image attention is windowed, while text/context tokens remain globally visible so that conditioning is not sparsified.

In practice, this can be implemented directly, or can be implemented by leveraging masking options in existing attention kernels, which gives a significant speedup.

\subsection{Conditioning and Locality Gap}
The conditioning and locality gap are defined in the main text. The training trajectory is generated by taking an image from the dataset, and evolving it forward by adding noise. We only compute the training trajectory for Facebook DiT since its training dataset is ImageNet which is publicly accessible~\cite{deng2009imagenet}. Also note that there is no distinction between conditional and unconditional training trajectory. The sampling trajectory is generated by starting from complete noise and evolving it backward using the diffusion model. Sampling trajectory can be either conditional or unconditional. We also note that all score gaps and their $l_2$ norms are directly computed in the latent space. We observe that score gaps in latent space translates to a larger gap in the pixel space when $t$ is closer to zero. This tilts the score gap plot towards the beginning. We hypothesize that this enhanced sensitivity is due to the formation of data manifold near clean image which could enhance perturbation in the latent space. We report the $l_2$ norm divided by the number of image tokens.

For Facebook DiT-XL results presented in Figure \ref{fig:score-gap}, we take $500$ steps along the training and sampling trajectory. We take $20$ samples at each time step. We use the HuggingFace diffusers implementation of DPMSolverMultistepScheduler~\cite{lu2022dpm,lu2025dpm}. We evaluate $20$ samples for each time step. Image size is fixed to $256 \times 256$ by the model. For conditioning, we use class $207$ (golden retriever). We do not observe qualitative differences between different class labels.

For SD3 medium, we take $100$ steps along the sampling trajectory. We take $10$ samples at each time step. We use the StableDiffusion3Pipeline from HuggingFace diffusers. We use an image size of $512 \times 512$. We use a CFG strength of 4.0 in both cases. For conditioning, we use prompt ``a golden retriever playing in a park, high detail, soft lighting''. Again, we do not observe qualitative differences between differnt prompts.

For both models, the VAE has a eight-to-one compression ratio, and they both organize a two-by-two latent patch into a token. Therefore, both model have a token patch size of $16 \times 16$ in the pixel space. This means that for Facebook DiT-XL, setting $r=16$ reduces local attention to global dense attention. For SD3 medium in our setup, this $r$ becomes $32$.

Subsequent experiments inherit the same setup unless otherwise noted.

\subsection{Forward-backward experiments}
For Facebook DiT, we use $20$ clean-generation steps, $50$ denoising steps, and evaluate at $40$ linearly spaced noise levels with $t_{\mathrm{norm}}\in[0.1,1.0]$. For each noise level, the experiment draws $40$ independent noise realizations. 

For both models, We use microsoft/resnet-50 for classification~\cite{he2016deep,resnet50_hf}. We report the standard deviation of the classifier error which is analytically determined from the Bernoulli distribution, since classification is either correct or incorrect. For a datapoint with classification error $p$ and with $n$ samples, the one-sigma error is given by $\sqrt{p(1-p)/n}$.

\subsection{Conditioning window and locality window experiments}
We use $40$ sampling steps for all window experiments. For conditioning window experiments for both models, we use a window size of $0.1$ and scan $t_i$ from $0$ to $0.9$ by steps of $0.1$ (setting $t_i > 0.9$ truncates the window size). We use $500$ samples per $t_i$. For locality window experiments for Facebook DiT, we use a window size of $0.4$ and scan $t_i$ from $0.2$ to $0.8$ by steps of $0.05$ (setting $t_i > 0.9$ or $t_i < 0.2$ truncates the window size). We use $500$ samples per $t_i$. For SD3, we start $t_i$ at $0.4$ and use $62$ samples per $t_i$ to reduce compute.

Our FID follows from the standard definition~\cite{heusel2017gans}. We use the pool3 layer of the inception V3 network~\cite{szegedy2016rethinking} as features in computing the Gaussian Fréchet distance. Note that we are in a sample-deficient regime which will lead to significant overestimate of FID. This is fine since we are not concerned about the absolute value of FID, but at which $t_i$ it becomes the smallest. We compute the one-sigma error by bootstrap resampling.

\section{Limitation of Our Work}\label{sec:Limitation}
Our empirical study is currently limited to two representative diffusion transformer systems: a class-conditional ImageNet DiT and a text-conditioned MMDiT/SD3 model.
Although these models cover two common conditioning mechanisms and show consistent concurrence between the symmetry-breaking and nonlocality critical windows, a broader evaluation across architectures, model scales, datasets, samplers, and prompt distributions is needed before claiming universality.
In addition, our truncated-attention DiT should be understood as an approximate local denoiser rather than a strictly local one, where information can still propagate across distant patches through multiple layers of local attention. 
Constructing a numerically faithful strictly local score estimator for modern pretrained diffusion models is itself challenging; future work could test whether the same concurrence persists under more diverse architectures and under stricter or independently trained local denoisers.

Our conclusions are also empirical rather than theoretical. The observed concurrence is supported by score-gap measurements and windowed sampling interventions, but we have not proven that symmetry-breaking and nonlocality transitions must coincide in general.
Moreover, our analysis has not yet systematically compared how the identified critical windows in the latent space.
Some quantitative estimates, especially FID and sliding-window results, are based on finite samples and should be interpreted as evidence for relative timing rather than precise estimates of the phase transition window. Developing sharper statistical tests and a theory connecting these operational probes to Markov length or conditional mutual information remains an important direction for future work.

We are also not able to quantify the width of the critical window. We expect that different definitions give rise to different width. Only in well-defined ``thermodynamic limit'' can one give a unified definition (and critical window can also collapse to a critical point in this limit). Ideally, it would be useful to understand the criticality better by probing certain universal scaling behavior as in statistical mechanics. However, it is currently unclear what is the correct scaling variable and scaling hypothesis that captures the behavior of diffusion models. We leave such a scaling analysis to future work.


\newpage

\end{document}